\documentclass[]{bytedance}
\usepackage[toc,page,header]{appendix}

\usepackage{minitoc}
\usepackage{amsfonts}
\usepackage{amssymb}
\usepackage{tabularx}
\usepackage{listings}
\usepackage{xcolor}
\usepackage{cancel}

\usepackage{tabulary,multirow,xspace}
\usepackage{fixmath,mathtools,nicefrac,mmstyle}
\usepackage{subcaption}
\usepackage{caption}
\usepackage{wrapfig} 
\usepackage[misc]{ifsym} 
\usepackage{colortbl}

\usepackage{wrapfig}
\usepackage{multicol}
\usepackage[most]{tcolorbox}
\usepackage{pifont}

\definecolor{codegreen}{rgb}{0,0.6,0}
\definecolor{codegray}{rgb}{0.5,0.5,0.5}
\definecolor{codepurple}{rgb}{0.58,0,0.82}
\definecolor{backcolour}{rgb}{0.95,0.95,0.92}
\definecolor{boxblue}{RGB}{57,89,163}
\definecolor{boxbluebg}{RGB}{230,237,250} 

\lstdefinestyle{mystyle}{
    backgroundcolor=\color{backcolour},   
    commentstyle=\color{codegreen},
    keywordstyle=\color{magenta},
    numberstyle=\tiny\color{codegray},
    stringstyle=\color{codepurple},
    basicstyle=\ttfamily\footnotesize,
    breakatwhitespace=false,         
    breaklines=true,                 
    captionpos=b,                    
    keepspaces=true,                 
    numbers=none,                    
    numbersep=5pt,                  
    showspaces=false,                
    showstringspaces=false,
    showtabs=false,                  
    tabsize=2
}
\definecolor{mygray1}{gray}{.95}
\definecolor{mygray2}{gray}{.9}
\definecolor{mygray3}{gray}{.95}
\usepackage{pifont}

\newlength\savewidth
\newcolumntype{x}[1]{>{\centering\arraybackslash}p{#1pt}}

\newcommand{\app}{\raise.17ex\hbox{$\scriptstyle\sim$}}

\makeatletter
\DeclareRobustCommand\onedot{\futurelet\@let@token\@onedot}
\def\@onedot{\ifx\@let@token.\else.\null\fi\xspace}

\makeatother

\makeatletter

\newcommand{\Rmnum}[1]{\expandafter\@slowromancap\romannumeral #1@}
\makeatother

\usepackage{xcolor}
\usepackage{graphicx}
\usepackage{amssymb}
\usepackage{pifont}
\usepackage{floatrow}
\usepackage{amsmath} 
\usepackage{float}
\usepackage{wrapfig}
\usepackage{multirow}
\usepackage{tcolorbox}
\tcbuselibrary{breakable, skins, raster}
\usepackage{listings}
\usepackage{listings}

\definecolor{commentgreen}{rgb}{0.1, 0.4, 0.1}
\definecolor{keywordblue}{rgb}{0.1, 0.1, 0.7}
\definecolor{stringred}{rgb}{0.7, 0.1, 0.1}

\lstdefinestyle{mystyle}{
    commentstyle=\color{commentgreen},
    keywordstyle=\color{keywordblue},   
    stringstyle=\color{stringred},
    basicstyle=\ttfamily\scriptsize, 
    breaklines=true,
    keepspaces=true,
    showstringspaces=false,
    frame=none,                     
    language=Python, 
}

\usepackage{xspace}
\usepackage{makecell}
\usepackage{booktabs}
\usepackage{amssymb} 
\usepackage{pifont} 
\usepackage{dblfloatfix}
\usepackage{enumitem}
\usepackage{multirow}
\usepackage{rotating}
\usepackage{array}
\usepackage[autostyle=true]{csquotes}

\usepackage{colortbl}
\usepackage{xcolor}
\definecolor{Red}{RGB}{192, 0, 0}
\definecolor{Blue}{RGB}{12, 114, 186}
\definecolor{Yellow}{RGB}{218, 169, 20}
\definecolor{lightyellow}{RGB}{255,255,153}

\definecolor{HighlightBlue}{RGB}{0, 100, 148}
\definecolor{HighlightRed}{RGB}{230, 57, 70}

\definecolor{LightRed}{HTML}{ffe0e0}
\definecolor{LightBlue}{HTML}{def5ff}
\definecolor{LightYellow}{HTML}{FFF6DB}
\definecolor{LightGreen}{HTML}{eff9f0}

\usepackage{hyperref}
\usepackage{url}
\usepackage{pifont}
\usepackage{soul} 
\usepackage{color, xcolor}

\usepackage{wrapfig,lipsum,booktabs}

\usepackage{colortbl}
\definecolor{lightyellow}{RGB}{255,242,204}
\definecolor{lightorange}{RGB}{251,229,214}
\definecolor{lightgreen}{RGB}{226,240,217}
\definecolor{lightblue}{RGB}{222,235,247}
\definecolor{lightgray}{RGB}{209,201,206}
\definecolor{deepgray}{RGB}{178,164,173}
\definecolor{deepblue}{RGB}{112,168,218}

\usepackage{threeparttable}
\usepackage{xspace}
\usepackage{graphicx} 
\usepackage{float} 
\usepackage{booktabs}
\usepackage{multicol}
\usepackage{multirow} 
\usepackage[ruled,boxed,linesnumbered]{algorithm2e}
\usepackage{caption}
\usepackage{cleveref}
\crefname{figure}{Fig.}{Figs.}  
\Crefname{figure}{Fig.}{Figs.}  
\crefname{table}{Tab.}{Tabs.}  
\Crefname{table}{Tab.}{Tabs.}   
\crefname{section}{Sec.}{Secs.} 
\Crefname{section}{Sec.}{Secs.}   
\crefname{appendix}{Appendix}{Appendix}   
\Crefname{appendix}{Appendix}{Appendix}

\usepackage{setspace}

\usepackage{tikz}

\usepackage{booktabs,threeparttable,tabularx,array,ragged2e,xcolor,siunitx}
\newcolumntype{L}[1]{>{\raggedright\arraybackslash}p{#1}}

\newtcolorbox{promptbox}[1][]{
  enhanced,
  breakable,
  colback=Blue!5!white,      
  colframe=Blue!75!black,    
  coltitle=white,
  fonttitle=\bfseries,
  title=#1,
  boxed title style={
    colback=Blue!75!black,   
    boxrule=0pt,
    arc=2pt,
    top=1mm,
    bottom=1mm,
    left=3mm,
    right=3mm
  },
  sharp corners,
  boxrule=0.5pt,
  before skip=10pt, after skip=10pt
}

\title{Bridging Your Imagination with Audio-Video Generation via a Unified Director}

\author{
\centerline{
    Jiaxu Zhang$^{1,2}$\quad 
    Tianshu Hu$^{1,{\dagger}}$ \quad  
    Yuan Zhang$^{1}$ \quad 
    Zenan Li$^{1}$ \quad
    \vspace{5pt}
} 
\centerline{
    Linjie Luo$^{1}$ \quad
    Guosheng Lin$^{2,{\ddagger}}$ \quad
    Xin Chen$^{1,{\dagger},{\ddagger}}$\quad
    \vspace{-5pt}
}
}

\affiliation[1]{ByteDance Intelligent Creation}
\affiliation[2]{Nanyang Technological University}
\contribution[\dagger]{Project Lead}
\contribution[\ddagger]{Corresponding Authors}

\abstract{
Existing AI-driven video creation systems typically treat script drafting and key-shot design as two disjoint tasks: the former relies on large language models, while the latter depends on image generation models. We argue that these two tasks should be unified within a single framework, as logical reasoning and imaginative thinking are both fundamental qualities of a film director. 
In this work, we propose UniMAGE, a unified director model that bridges user prompts with well-structured scripts, thereby empowering non-experts to produce long-context, multi-shot films by leveraging existing audio–video generation models.
To achieve this, we employ the Mixture-of-Transformers architecture that unifies text and image generation. To further enhance narrative logic and keyframe consistency, we introduce a ``first interleaving, then disentangling'' training paradigm. Specifically, we first perform Interleaved Concept Learning, which utilizes interleaved text–image data to foster the model’s deeper understanding and imaginative interpretation of scripts. We then conduct Disentangled Expert Learning, which decouples script writing from keyframe generation, enabling greater flexibility and creativity in storytelling. 
Extensive experiments demonstrate that UniMAGE achieves state-of-the-art performance among open-source models, generating logically coherent video scripts and visually consistent keyframe images.
}

\date{\today}

\checkdata[Project Page]{\url{https://kebii.github.io/UniMAGE}}

\begin{document}
\maketitle

\section{Introduction}

\vspace{1mm}

\begin{quote}
\textit{"A writer needs a pen, an artist needs a brush, but a filmmaker needs an army."}
\hfill --- Orson Welles
\end{quote}

\vspace{1mm}

\begin{figure}[h]
\begin{center}
   \includegraphics[width=1.0\linewidth]{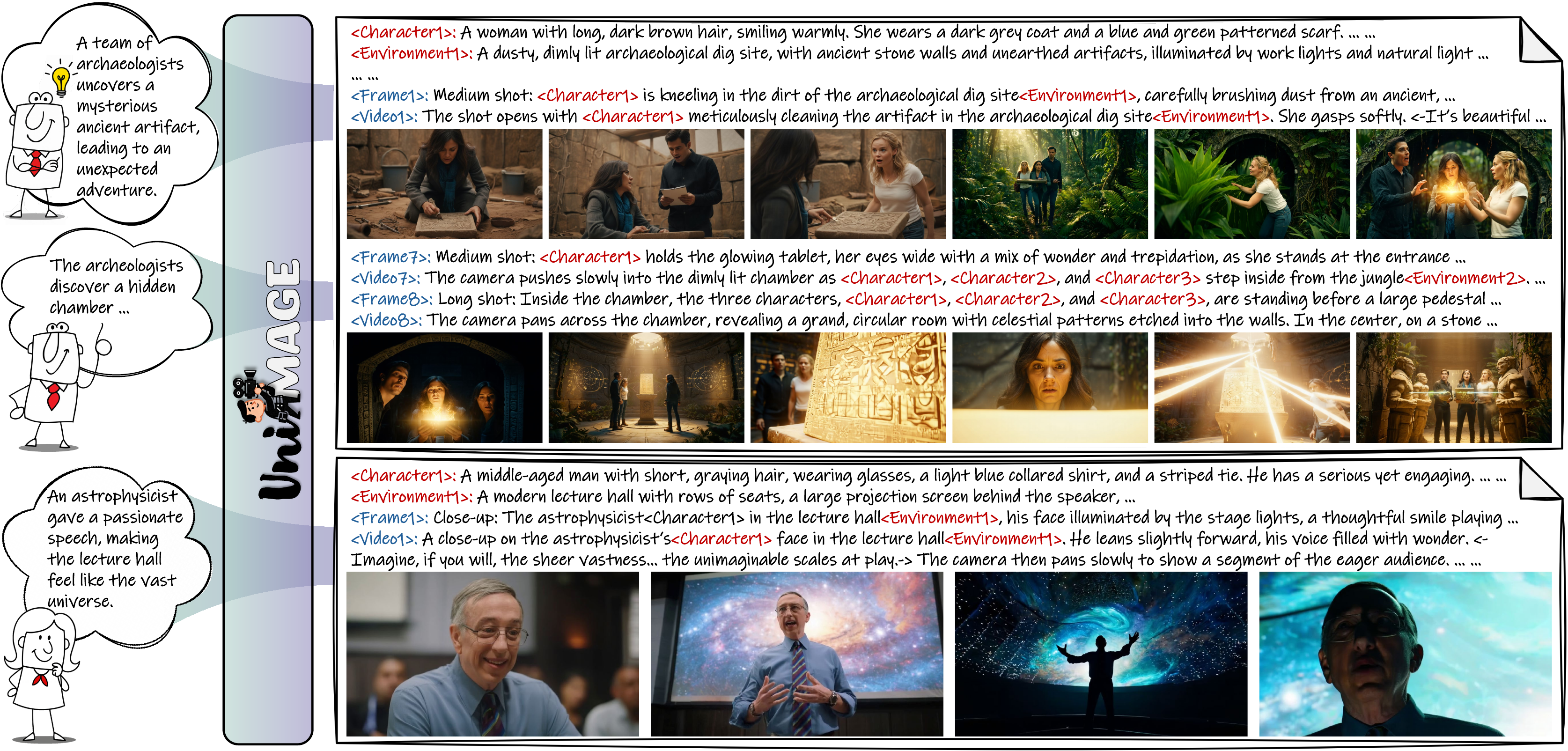}
\end{center}
   \caption{\textbf{Showcase of UniMAGE’s multimodal directing abilities.} UniMAGE unifies script drafting, extension, continuation, and keyframe image generation, thereby enabling coherent long-form storytelling with consistent characters and cinematic visual compositions. The generated scripts and keyframes can further serve as structured, high-level guidance for existing audio-video joint generation models.}
\label{fig:teaser}
\end{figure}

Audio-video generation has advanced rapidly in recent years, with models such as Veo 3~\citep{veo3}, Sora 2~\citep{sora2}, and Keling 2.5~\citep{klingai} demonstrating remarkable creative potential and inspiring users across online communities. However, these models primarily focus on visual fidelity and temporal coherence within short, single-shot videos, which limits their ability to convey long-form narratives. To address this limitation, recent approaches~\citep{long2024videostudio, wu2025automated, huang2025filmaster} have begun to employ large language models (LLMs) and image generation models as collaborative agents that align user prompts with multi-shot video captions and generate key images for each shot, thereby facilitating film-level video generation. Nevertheless, these methods typically treat script drafting and keyframe generation as two disjoint tasks, resulting in limited narrative logic and visual consistency.

In traditional filmmaking, scriptwriting and storyboard design are inseparable processes that require an experienced director to systematically plan both the overall narrative and the visual composition. A film director functions much like a conductor in an orchestra, bringing out the strengths of each element while maintaining overall harmony. Inspired by this analogy, we posit that a unified understanding and generation model is essential to bridge user intents with audio-video generation systems, rather than relying on separate agents. Accordingly, in this work, we present a unified director model named UniMAGE.

UniMAGE is built upon the Mixture-of-Transformers (MoT) architecture proposed in Bagel~\citep{deng2025emerging}, which integrates an LLM and a diffusion model within a single transformer framework, enabling joint reasoning and generative learning across text and image modalities. Although MoT provides a strong, unified architecture, UniMAGE faces a more challenging task than previous multimodal systems, as film scripts inherently involve lengthy, context-rich sequences.

On the one hand, maintaining image consistency across long-context sequences is extremely challenging. Previous studies~\citep{zhou2024storydiffusion, dinkevich2025story2board, xie2024show} mainly focused on preserving a single subject across frames or conducting limited-step image editing. However, when scripts involve multiple actors, existing models struggle to reliably recognize and maintain individual identities. UniMAGE addresses this issue through \textit{Interleaved Concept Learning}. Specifically, we organize the script and corresponding keyframes into a text–image interleaved format and jointly optimize the parameters of the LLM and diffusion transformer. This interleaved representation enables the model to better capture the overall narrative concept. To further enhance image consistency, we introduce \textit{In-Context ID Prompting} strategy, which inserts special tokens among the reference ViT and VAE tokens to indicate the indices of generated keyframes and the identities of the characters, thereby enabling stable multi-subject appearances across long sequences.

On the other hand, ensuring coherent script logic and supporting flexible continuation remain highly demanding. Previous studies~\citep{huang2025filmaster, wu2025automated, long2024videostudio} mainly relied on prompting a pre-trained LLM to generate scripts, which requires per-sample prompt engineering and often overlooks reasonable storyboard design, making it difficult to preserve logical coherence and temporal alignment in multi-scene storytelling. UniMAGE overcomes this limitation through \textit{Disentangled Expert Learning}. Specifically, we maintain keyframe generation in an interleaved format but freeze the LLM parameters when optimizing the diffusion loss, and decouple the script-writing process by training the LLM as a dedicated script expert. To further support flexible script continuation, we employ the \textit{Pre-Context Script Splitting} strategy, which randomly divides complete scripts so that the model learns to continue script generation based on the preceding context or a given user prompt, thereby ensuring logically coherent and adaptable narrative development.

Our contributions are summarized as follows:

\noindent\textbf{Concept:} UniMAGE embodies the concept of a ``unified director model" that holistically orchestrates narrative logic and visual composition, bridging user intent with multimodal script for creative audio-video generation.

\noindent\textbf{Technique:} UniMAGE employs Interleaved Concept Learning and Disentangled Expert Learning with In-Context ID Prompting and Pre-Context Script Splitting to enhance visual consistency and narrative coherence over long-context sequences.

\noindent\textbf{Performance:} UniMAGE demonstrates strong capabilities and generalization in long-form, multi-scene script creation, achieving superior narrative coherence, character stability, and image consistency compared with existing agent-based and unified models.
\section{Related Work}

\noindent\textbf{Generative Models for Film Production.} Recent studies~\citep{huang2025filmaster, xiao2025captain, shi2025animaker, cheng2024theatergen, mao2024story_adapter} have leveraged the emerging reasoning and planning capabilities of large language models, together with the image generation capabilities of diffusion models, to advance AI-assisted filmmaking. For script creation, Anim-Director~\citep{li2024anim} employs an LLM to expand high-level prompts into detailed and structured scripts. FilmAgent~\citep{xu2025filmagent} simulates a director’s workflow in 3D virtual environments, covering idea development, scriptwriting, and cinematography to support the end-to-end filmmaking process. MovieAgent~\citep{wu2025automated} adopts a multi-agent Chain-of-Thought strategy to automate script breakdown, scene planning, and shot design.

For keyframe generation, StoryDiffusion~\citep{zhou2024storydiffusion} introduced a plug-and-play attention module for diffusion models to maintain character consistency across image sequences. Story2Board~\citep{dinkevich2025story2board} used a lightweight consistency mechanism to preserve the model’s generative prior, enabling expressive storytelling. Furthermore, FilMaster~\citep{huang2025filmaster} is a comprehensive AI-based film generation system explicitly designed around cinematic principles to guide camera language and rhythm. However, most of these systems decoupled script generation and visual synthesis into separate agents, leading to weak narrative–visual alignment and limited long-range coherence. Although SEED-Story~\citep{yang2025seed} leverages a unified MLLM to produce multimodal stories, it requires separate training for each story instance, and its generated images still suffer from limited visual quality. In contrast, UniMAGE integrates script reasoning and generation within a mixed transformer architecture, enabling end-to-end generalizable learning of both narrative logic and high-quality visual composition in a single framework.

\vspace{1mm}
\noindent\textbf{Unified Multimodal Generation.} Recent efforts have explored two main approaches to unifying multimodal understanding and visual generation. One adopts an auto-regressive (AR) framework to jointly produce text and image tokens, as in Emu~\citep{wang2024emu3} and Chameleon~\citep{team2024chameleon}, which treat images as continuous or discrete visual tokens alongside text. However, AR models generally generate lower-quality images compared with diffusion-based approaches. The other line of research integrates diffusion modules into large language models, such as Show-O~\citep{xie2024show} and TransFusion~\citep{zhou2024transfusion}, which synthesize text and images within a shared transformer backbone. Bagel~\citep{deng2025emerging} further introduces a Mixture-of-Transformers architecture that selectively activates modality-specific parameters while enabling long-context interaction between multimodal understanding and generation through shared self-attention mechanisms. Nevertheless, current unified models mainly handle short-context reasoning and generation, without mechanisms to preserve narrative or visual coherence across long sequences. Based on the MoT framework, our UniMAGE jointly learns narrative structure and visual composition, enabling coherent long-form script and keyframe generation.

\vspace{1mm}
\noindent\textbf{Audio-Video Generation.} Early studies on joint audio-video generation, such as MM-Diffusion~\citep{ruan2023mm}, used U-Net architectures to model audio and video separately. Later works like SyncFlow~\citep{liu2024syncflow}, Uniform~\citep{zhao2025uniform}, and UniVerse-1~\citep{wang2025universe} expanded training on large datasets, improving cross-modal alignment and generalization. Recent commercial systems, including Veo 3~\citep{veo3} and Sora 2~\citep{sora2}, have achieved notable progress in synchronized audio–video generation. However, these models rely heavily on well-structured prompts and are typically limited to short, single-shot videos. To compensate, they often incorporate LLM-based re-captioning modules to refine user input before generation~\citep{gao2025seedance}. Despite these advances, current methods still lack the capability to plan and maintain coherence across multi-shot, narrative-driven sequences. To overcome this, we present UniMAGE, a unified director model that bridges user imagination with long-context, film-like audio–video creation through multimodal script generation.
\section{Method}
We introduce UniMAGE, a unified director model that transforms simple user prompts into illustrated scripts. Given a user prompt $\rho$, the overall pipeline of UniMAGE can be formulated as:
\begin{equation}
\text{UniMAGE}(\rho, \hat S) \mapsto (\mathcal{G}, \mathcal{C}, \mathcal{F}),
\end{equation}
where $\mathcal{G}$ denotes the global descriptions, such as character and environment definitions; $\mathcal{C}=(c_1,c_2,...,c_n)$ represents the content descriptions for $n$ video shots, including keyframe and video-level narratives; and $\mathcal{F}=(f_1,f_2,...,f_n)$ corresponds to the keyframe images for those shots. The full script is defined as $S=(\mathcal{G}, \mathcal{C}, \mathcal{F})$, and the superscript $\hat{S}$ indicates that the input is optional. $\hat{S}$ is used for script extension and continuation. Through this formulation, UniMAGE uniformly performs script creation, prompt-based or in-context script continuation, and keyframe generation, thereby empowering automatic video–audio creation.

UniMAGE adopts the MoT architecture, consisting of two transformer experts: one for multimodal understanding and another for image generation. Correspondingly, it uses two types of visual encoders, an understanding-oriented encoder (ViT), and a generation-oriented encoder (VAE). Both transformer experts process the same token sequence through shared self-attention layers. For text token prediction, UniMAGE follows the Next Token Prediction paradigm \cite{minaee2024large}, leveraging the established advantages of autoregressive modeling. For visual token prediction, it utilizes the Rectified Flow \cite{esser2024scaling}, aligning with prevailing practices in visual generation.

\subsection{Preliminaries} \label{sec:pre}

\noindent\textbf{Next Token Prediction (NTP)} serves as the fundamental training objective for auto-regressive models \cite{minaee2024large}. Given a token sequence $\mathbf{y}= (y_1, y_2, ..., y_n)$, where each token $y_t$ is drawn from a fixed vocabulary $V$, and assuming $\mathbf{y}$ follows a data distribution $P(y)$, the auto-regressive model $P_{\theta}$ decomposes the joint probability as: $P_{\theta}(\mathbf{y}) = \prod_{t=1}^{T} p_{\theta}(y_t \mid \mathbf{y}_{<t})$, where $\mathbf{y}{<t}= (y_1, y_2, ..., y_{t-1})$ denotes the sequence of tokens preceding position $t$. The model parameters $\theta$ are optimized to maximize the likelihood of the true token at each step, conditioned on its ground-truth history from the training dataset. Formally, the maximum likelihood estimation of $\theta$ can be expressed as:
\begin{equation} \label{Eq.1}
    \theta^{*} = \arg\max_{\theta} \sum_{t=1}^{T} \log p_{\theta}\left( y_{t} \mid \mathbf{y}_{<t} \right).
\end{equation}
During inference, the model generates tokens sequentially by either sampling or selecting the most probable token, conditioned on a given or learned context.

\vspace{1mm}
\noindent\textbf{Rectified Flow} is an ordinary differential equation (ODE) model that transports samples from $\pi_0$ to $\pi_1$ along straight trajectories \cite{liu2022flow}. Given two samples $X_{0} \sim \pi_0$ and $X_{1} \sim \pi_1$, the drift field $v$ is trained to approximate their difference through the following objective:
\begin{equation}
\min_{\phi} \int_{0}^{1} \mathbb{E} \left[ \| (X_{1} - X_{0}) - v_{\phi}(X_{t}, t) \|^{2} \right] dt,
\end{equation}
where $X_{t} = (1 - t)X_{0} + tX_{1}$. In practice, $v_{\phi}$ is parameterized by a neural network. After training, samples from $\pi_1$ can be obtained by integrating the ODE along the learned straight flow with only a few steps.

\begin{figure}[t]
\begin{center}
   \includegraphics[width=0.9\linewidth]{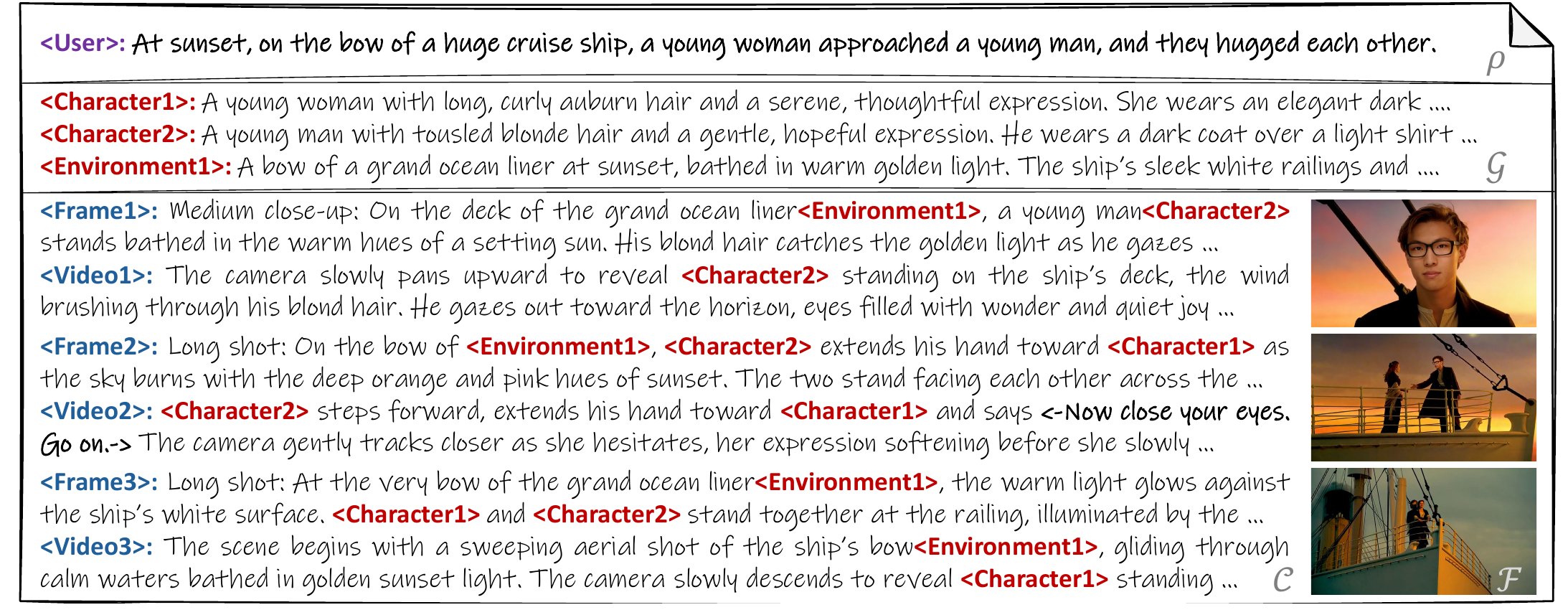}
\end{center}
   \caption{\textbf{Script structure of UniMAGE}. The script structure includes three components: global descriptions ($\mathcal{G}$), content descriptions ($\mathcal{C}$), and keyframe images ($\mathcal{F}$), together with a user prompt ($\rho$). Special tokens and indicator symbols are used to denote key script elements.}
\label{fig:script}
\end{figure}

\subsection{Script Structure} \label{sec:str}

As illustrated in Figure \ref{fig:script}, the script structure used in UniMAGE consists of three components: global descriptions $\mathcal{G}$, content descriptions $\mathcal{C}$, and keyframe images $\mathcal{F}$, along with a user prompt $\rho$. We design a set of special tokens to represent each element in the script, including $<$User$>$, $<$Character$N$$>$, $<$Environment$N$$>$, $<$Frame$N$$>$, and $<$Video$N$$>$, where $N$ denotes the index. When a character or environment defined in the global description appears within the content description, the corresponding special token is used to indicate the subject’s occurrence in the scene—for example, ``On the deck of the grand ocean liner$<$Environment1$>$, a young man$<$Character2$>$ …''. These special tokens help the model accurately identify characters and environments, while keeping the script concise and structurally consistent.

The content descriptions are further divided into two complementary layers. Frame descriptions capture the static visual layout of key moments, such as camera position, lighting, and character placement. Video descriptions, in contrast, focus on temporal and narrative aspects, including dialogues, plot progression, and actions. In addition, we introduce indicator symbols $<$- -$>$ to denote character dialogue and environmental sound effects, allowing the corresponding audio content to be easily retrieved in subsequent stages—for example, “$<$-Now close your eyes. Go on.-$>$”. To accommodate diverse user input formats, we define four distinct styles of user prompts, which are randomly sampled during training. Detailed definitions of these prompt styles are provided in the supplementary materials.

\subsection{Interleaved Concept Learning} \label{sec:int}
We initialize the MoT model of UniMAGE using the pre-trained weights from Bagel, which provide a strong foundational capability for unified multimodal understanding and generation. However, unlike Bagel’s training format, which focuses on multi-step image editing, UniMAGE is required to generate not only images but also script text conditioned on the understanding of preceding narratives. Moreover, the long-context nature of script data further exceeds the base model’s capacity, necessitating \textit{Interleaved Concept Learning} strategy for coherent narrative and visual generation.

\begin{figure}[t]
\begin{center}
   \includegraphics[width=1.0\linewidth]{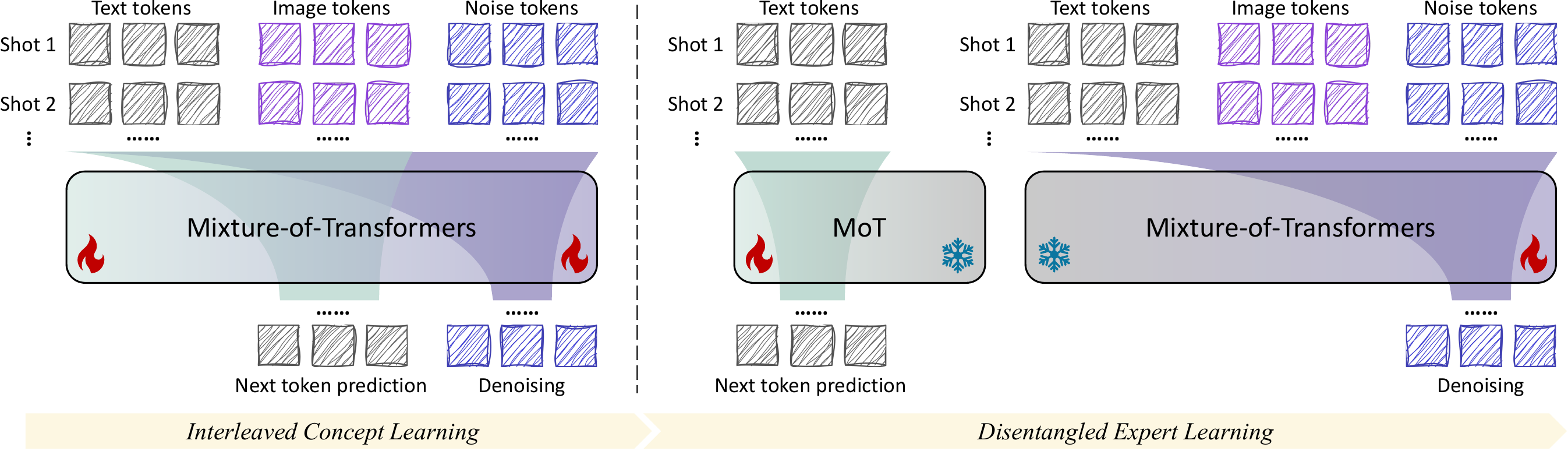}
\end{center}
   \caption{\textbf{Illustrations of Interleaved Concept Learning and Disentangled Expert Learning.} To enhance visual consistency and logical coherence across long-context scripts, as well as to fully leverage both textual and image data, we first optimize all MoT parameters using interleaved text–image data, and then disentangle the training of the understanding and generation experts—using pure text scripts for the former and text–image data for the latter.}
\label{fig:method1}
\end{figure}

With the script structure defined above, the script can be organized as interleaved text–image data. As illustrated in the left part of Figure \ref{fig:method1}, we first perform Interleaved Concept Learning, which enables the MoT model to generate text and images in an interleaved manner, thereby facilitating a deeper understanding of lengthy, context-rich scripts. This training stage is conceptually similar to the Chain-of-Thought strategy \cite{wei2022chain, fang2025got}, where the text content functions as the model’s reasoning process, followed by image generation conditioned on the preceding narrative context. In this stage, all parameters of the two transformer experts are jointly optimized, allowing the generated results to influence the model’s textual understanding and vice versa. Nevertheless, scripts often involve multiple characters and scenes, making it highly challenging for the model to consistently maintain the identities and visual coherence of different entities across long sequences. To address this, we propose \textit{In-Context ID Prompting} as follows.

\vspace{1mm}
\noindent\textbf{In-Context ID Prompting}. The core of addressing the visual consistency problem lies in enabling the model to recognize the characters and scenes depicted in each image and associate them with the global text descriptions and historical keyframes. Given that text tokens and image tokens are aligned within the pre-trained unified architecture, we can leverage the text to prompt and highlight key information represented in the images, thereby facilitating long-form visual consistency. As illustrated in the left part of Figure \ref{fig:method2}, within the ViT tokens used for understanding and the VAE tokens used for image reference, we insert special text tokens to indicate the frame ID, as well as the character and environment IDs appearing in each image. We apply full attention between each image’s ViT or VAE tokens and its corresponding special tokens. This In-Context ID Prompting strategy, together with the special tokens defined in our script structure, effectively preserves long-range associations between the script text and the generated images, ensuring consistent visual identity and scene continuity throughout the narrative.

\subsection{Disentangled Expert Learning} \label{sec:dis}

The Interleaved Concept Learning stage equips UniMAGE with a holistic and coherent understanding of the overall script. However, this interleaved generation strategy inevitably constrains the model’s flexibility in content creation, particularly for tasks such as script extension and continuation, where the model must dynamically adapt to new user prompts or seamlessly expand upon existing narrative contexts. In addition, obtaining logically consistent multi-shot text-image data is inherently difficult, which limits the model’s ability to fully learn long-form narrative logic from interleaved data. Consequently, we introduce \textit{Disentangled Expert Learning} strategy.

As illustrated in the right part of Figure \ref{fig:method1}, in this training stage, we decouple script content generation from interleaved keyframe generation and optimize the understanding transformer expert using pure text scripts. Meanwhile, the generation transformer expert is further optimized using interleaved text–image data, with the understanding branch frozen via stop-gradient. In addition, we incorporate text–image pairs into the training process to further improve visual fidelity. Through this strategy, both script logic and image quality are effectively enhanced, as the model can fully exploit heterogeneous multimodal training data beyond the interleaved script data. Finally, to enable script extension and continuation, we introduce the \textit{Pre-Context Script Splitting} strategy, described as follows.

\begin{figure}[t]
\begin{center}
   \includegraphics[width=1.0\linewidth]{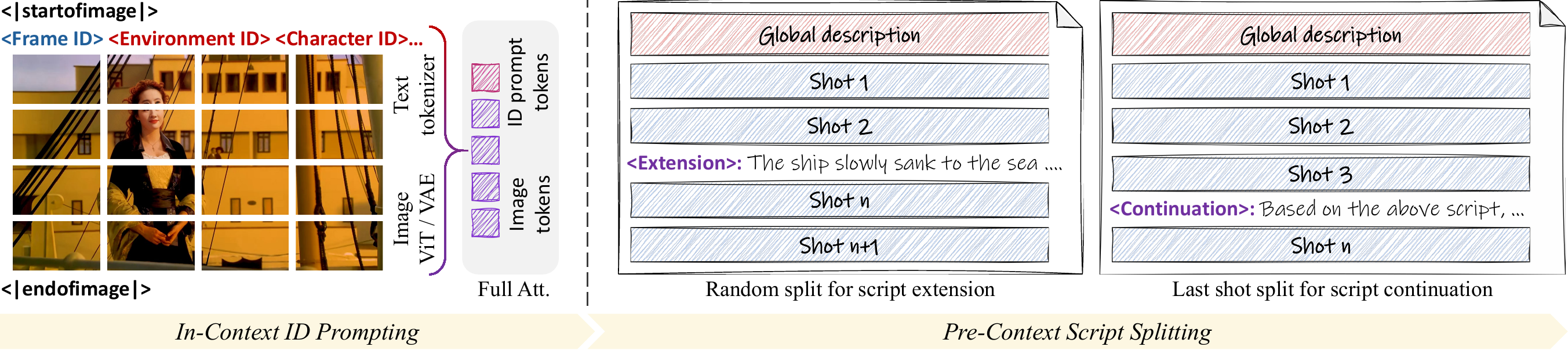}
\end{center}
   \caption{\textbf{Illustrations of In-Context ID Prompting and Pre-Context Script Splitting.} The former enhances visual consistency by aligning generated images with global character and scene descriptions, while the latter enables adaptive narrative extension and continuation.}
\label{fig:method2}
\end{figure}

\vspace{1mm}
\noindent\textbf{Pre-Context Script Splitting}. Based on the pure text scripts, we randomly insert new user or system prompts to simulate two types of creative demands. The first is prompt-based script extension, as illustrated in the middle part of Figure \ref{fig:method2}. Specifically, we divide a complete script into two parts and insert the indicator token $<$Extension$>$, followed by a new user prompt within the script, allowing the model to learn how to extend an existing narrative coherently from a given prompt. The new user prompt is generated by summarizing the second part of the script using Qwen 2.5 \cite{hui2024qwen2}. The second is in-context script continuation, as shown in the right part of Figure \ref{fig:method2}. In this case, we insert the indicator token $<$Continuation$>$, followed by a system prompt before the last shot of the script, enabling the model to infinitely continue the script during inference.

\subsection{Inference of UniMAGE} \label{sec:inf}

During inference, we maintain the disentangled generation process for text and images. Specifically, UniMAGE first generates a multi-shot text script conditioned on the user prompt. The user can then extend the narrative with new prompts or continuously generate subsequent shots based on the previously generated content. Finally, the complete script is segmented into individual shots, and corresponding keyframe images are generated in an interleaved manner. This unified yet disentangled strategy—where a single model handles both modalities while separating the generation processes for text and image—effectively ensures the logical coherence of the script and the visual consistency of the generated images. As a result, UniMAGE can produce longer, more coherent narratives while mitigating issues such as plot repetition and image distortion.
\begin{figure}[t]
\begin{center}
   \includegraphics[width=1.0\linewidth]{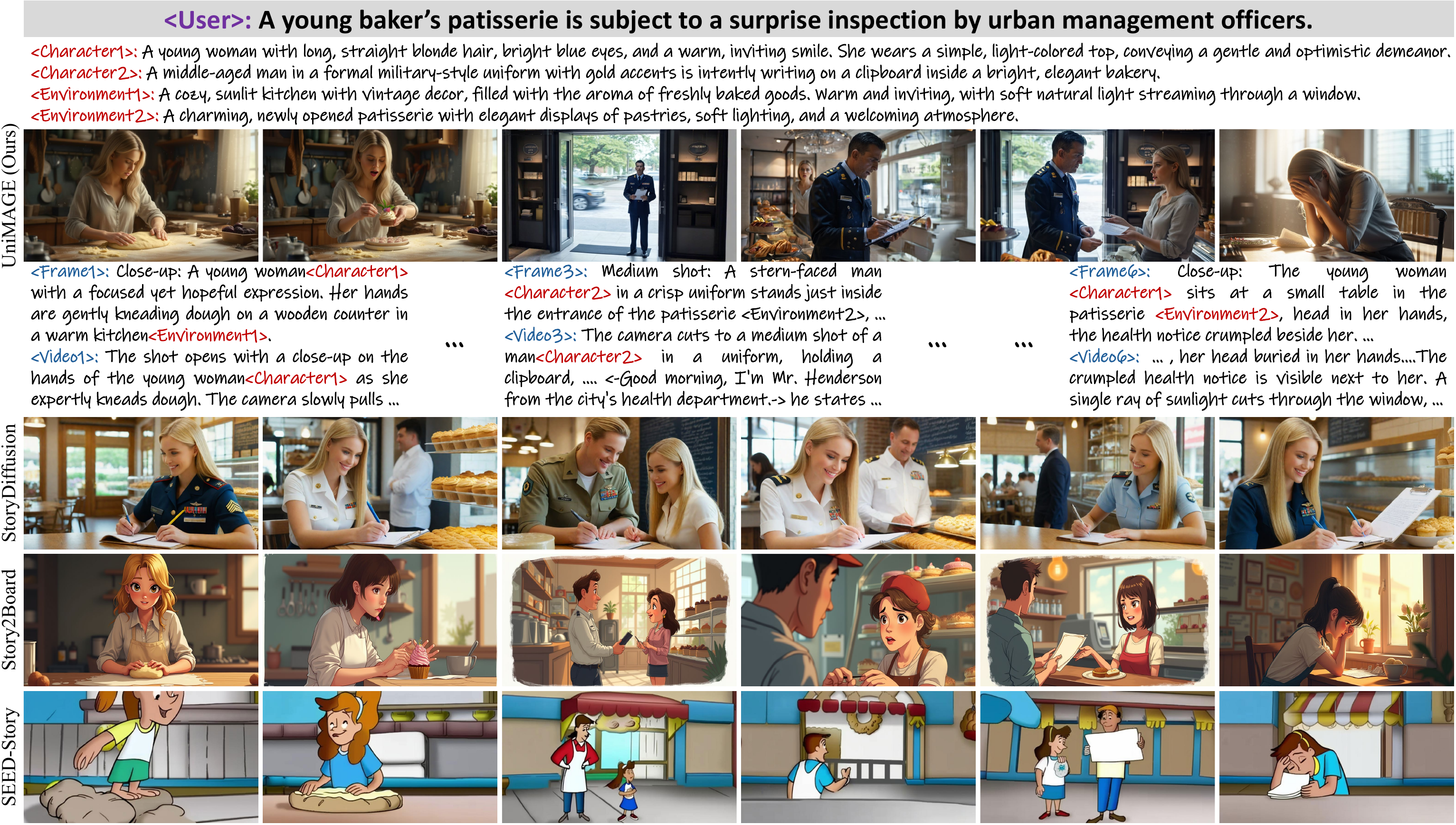}
\end{center}
   \caption{\textbf{Comparison with the baselines for multi-character script generation.} UniMAGE demonstrates a superior ability to maintain consistent character identities and visual coherence across multiple shots, whereas the baseline methods fail to preserve such consistency.}
\label{fig:exp_sota1}
\end{figure}

\section{Experiments}
\noindent\textbf{Datasets.} To support UniMAGE's unified multimodal training paradigm, we construct a large-scale and diverse dataset that integrates multi-shot scripts, long-form textual narratives, and high-quality text–image pairs. The dataset is composed of three complementary subsets, each aligned with a specific learning objective within UniMAGE:

(1) Multi-shot text–image scripts (450k sequences). We collect multi-shot videos from a broad range of open-source cinematic content, short films, and documentaries. Each video is segmented into coherent shots using visual scene transition detection. For each shot, we employ Gemini 2.5 Pro~\citep{Gemini} to generate detailed textual annotations. This subset forms the backbone for Interleaved Concept Learning, enabling UniMAGE to model multimodal reasoning and maintain global consistency across text–image interleaved sequences.

(2) Multi-shot text scripts (250k samples). To further enhance long-form narrative capability beyond visually grounded data, we curate a large corpus of textual scripts. These scripts are reorganized and structured using Qwen 2.5~\citep{hui2024qwen2} to fit the hierarchical script format of UniMAGE. This purely textual subset is essential for Disentangled Expert Learning, enabling the understanding expert to learn rich narrative logic, shot transitions, and dialog conventions.

(3) Single-shot text–image pairs (250k samples). To improve image quality and fidelity, particularly for character rendering and scene composition, we curate a large set of single-shot images. Each image is re-captioned with Gemini 2.5 Pro~\citep{Gemini} to obtain detailed, script-structured descriptions. This subset is leveraged in the generation expert’s training during the Disentangled Expert Learning stage, enabling improved visual precision, diversity, and controllability.

\vspace{1mm}
\noindent\textbf{Implementation details.} We implement UniMAGE based on the open-source framework of BAGEL~\citep{deng2025emerging}, which provides a unified MoT architecture as the foundation for multimodal understanding and generation. All experiments follow the standardized training pipeline and parallel strategy with BAGEL. During the Interleaved Concept Learning stage, only the multi-shot text–image script data is utilized, with a learning rate of 1e-5 and a total of 30,000 training steps. In the subsequent Disentangled Expert Learning stage, the entire dataset is employed, using the same learning rate of 1e-5 for 10,000 training steps. Notably, in this stage, only pure textual samples are used to optimize the understanding branch, while during the optimization of the generation branch, the text tokens are detached from the computational graph to prevent gradient propagation and ensure disentangled learning.

\vspace{1mm}
\noindent\textbf{Evaluation metrics.} We conduct both qualitative and quantitative evaluations to comprehensively assess the performance of UniMAGE across narrative coherence, character consistency, and visual quality. Qualitative results focus on long-form storytelling scenarios, where we present multi-shot scripts and corresponding keyframes generated from diverse user prompts. Quantitative results are obtained on the public benchmark ViStoryBench~\citep{zhuang2025vistorybench}, which assesses story visualization models across various narrative structures, visual styles, and character settings. We report six metrics: Style Similarity (CSD), Character Identification Similarity (CIDS), Prompt Adherence (Alignment), Onstage Character Count Matching (OCCM), Image Quality (Inception), and Aesthetics.

\begin{figure}[t]
\begin{center}
   \includegraphics[width=1.0\linewidth]{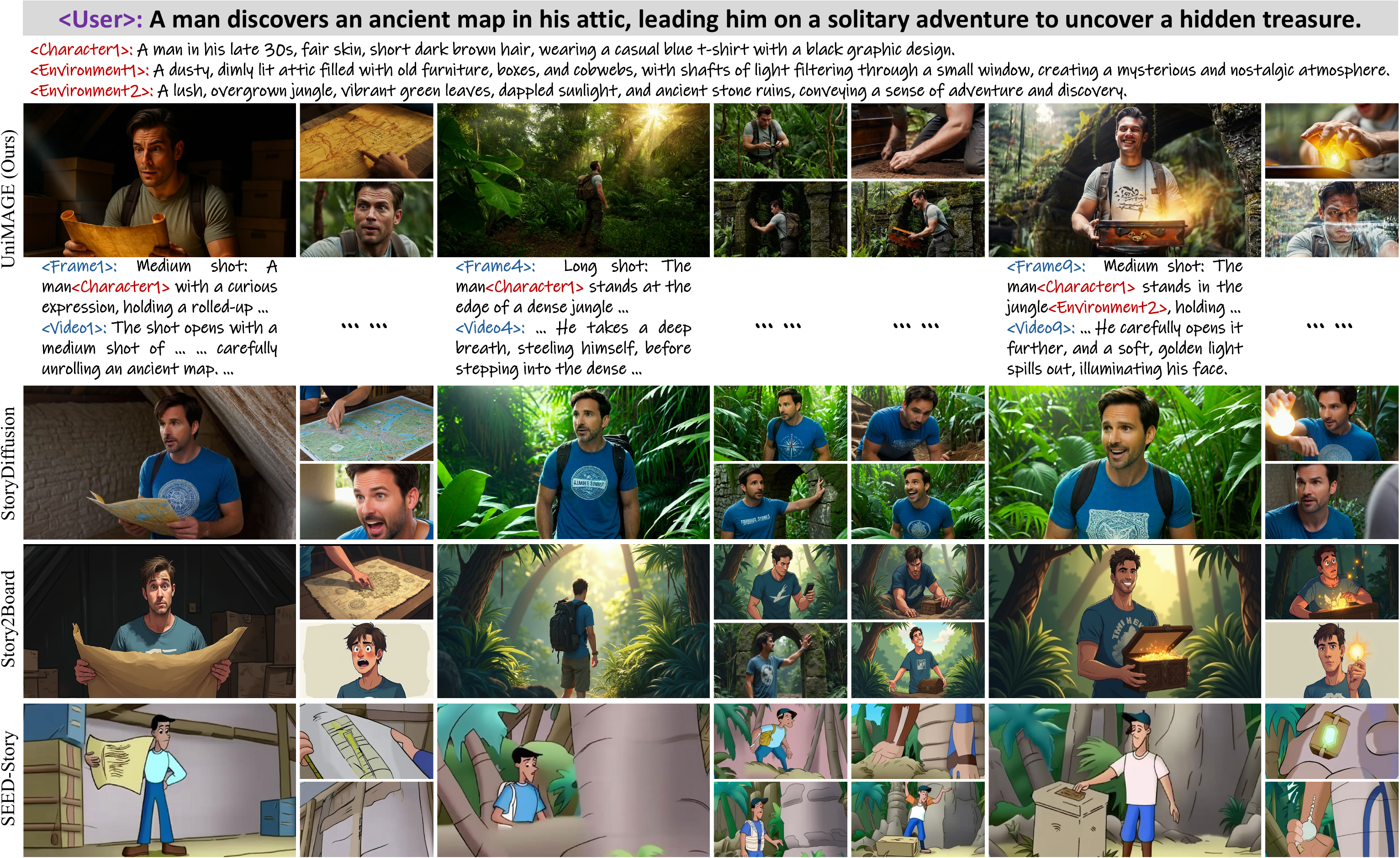}
\end{center}
   \caption{\textbf{Comparison with the baselines for long-form script generation.} UniMAGE demonstrates substantially improved visual coherence, narrative consistency, and scene diversity across long story sequences, whereas the baseline methods are limited in these aspects.}
\label{fig:exp_sota2}
\end{figure}

\begin{figure}[t]
\begin{center}
   \includegraphics[width=1.0\linewidth]{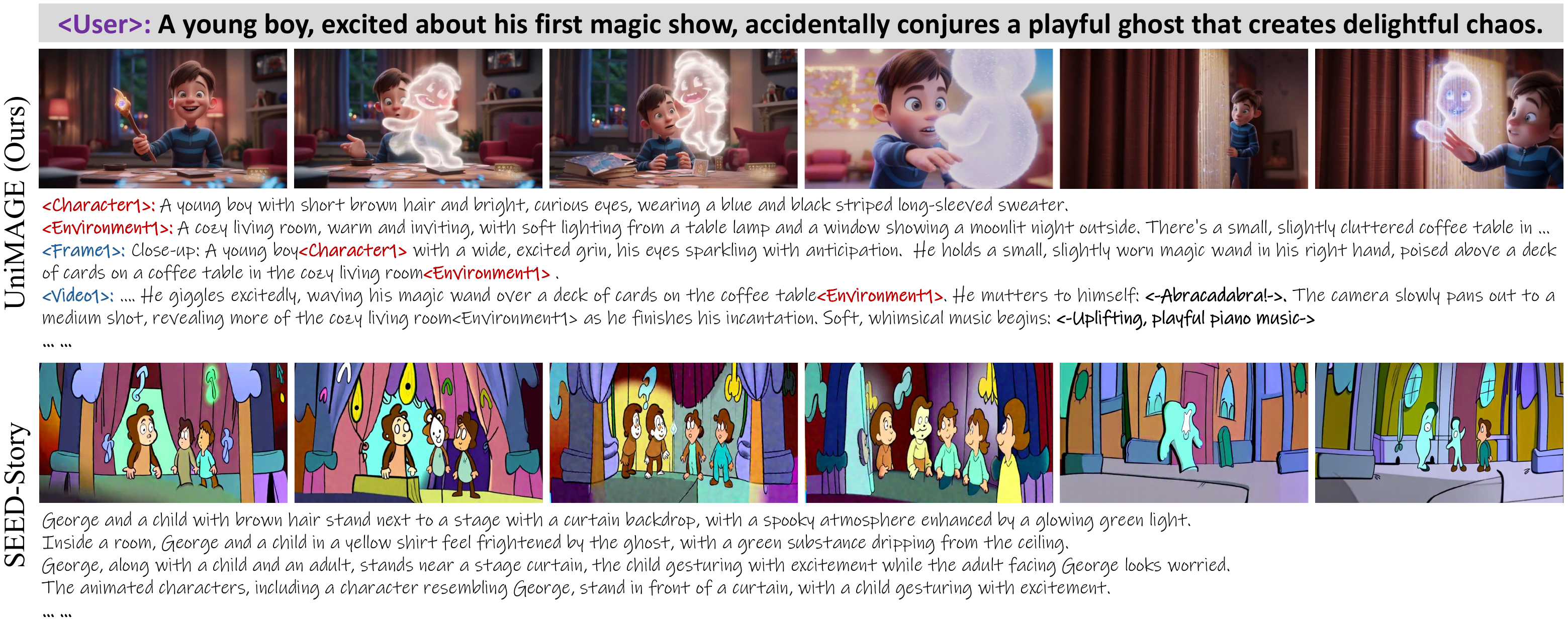}
\end{center}
   \caption{\textbf{Comparison with SEED-Story on multimodal script generation.} UniMAGE demonstrates superior logical coherence, visual quality, and cross-domain generalization, reflecting its stronger capability in unified multimodal understanding and generation.}
\label{fig:exp_seed}
\end{figure}

\subsection{Qualitative Results}
We compare UniMAGE with recent script visualization methods, including StoryDiffusion~\citep{zhou2024storydiffusion} and Story2Board~\citep{dinkevich2025story2board}, as well as the multimodal script generation model SEED-Story~\citep{yang2025seed}. The textual scripts used for StoryDiffusion and Story2Board are generated by UniMAGE. More results are presented in the supplementary materials.
 
 \vspace{1mm}
\noindent\textbf{Multi-character script generation.} Figure~\ref{fig:exp_sota1} compares UniMAGE with baseline models on multi-character script generation. While existing methods can generate plausible single frames, they generally struggle to maintain consistent identities across multiple shots. When the narrative shifts to new scenes or camera angles, baseline models often produce noticeable variations in facial structure, hairstyle, or clothing, leading to unstable and mismatched character appearances. In contrast, UniMAGE achieves stable identity preservation throughout the entire sequence. Thanks to its unified director architecture and the proposed In-Context ID Prompting, the model can reliably associate each character in the image with the corresponding textual identity defined in the script. This enables UniMAGE to maintain coherent visual traits even as the story spans different environments and shot configurations, resulting in noticeably more consistent and faithful multi-character narratives.

\begin{figure}[!htbp]
\begin{center}
   \includegraphics[width=1.0\linewidth]{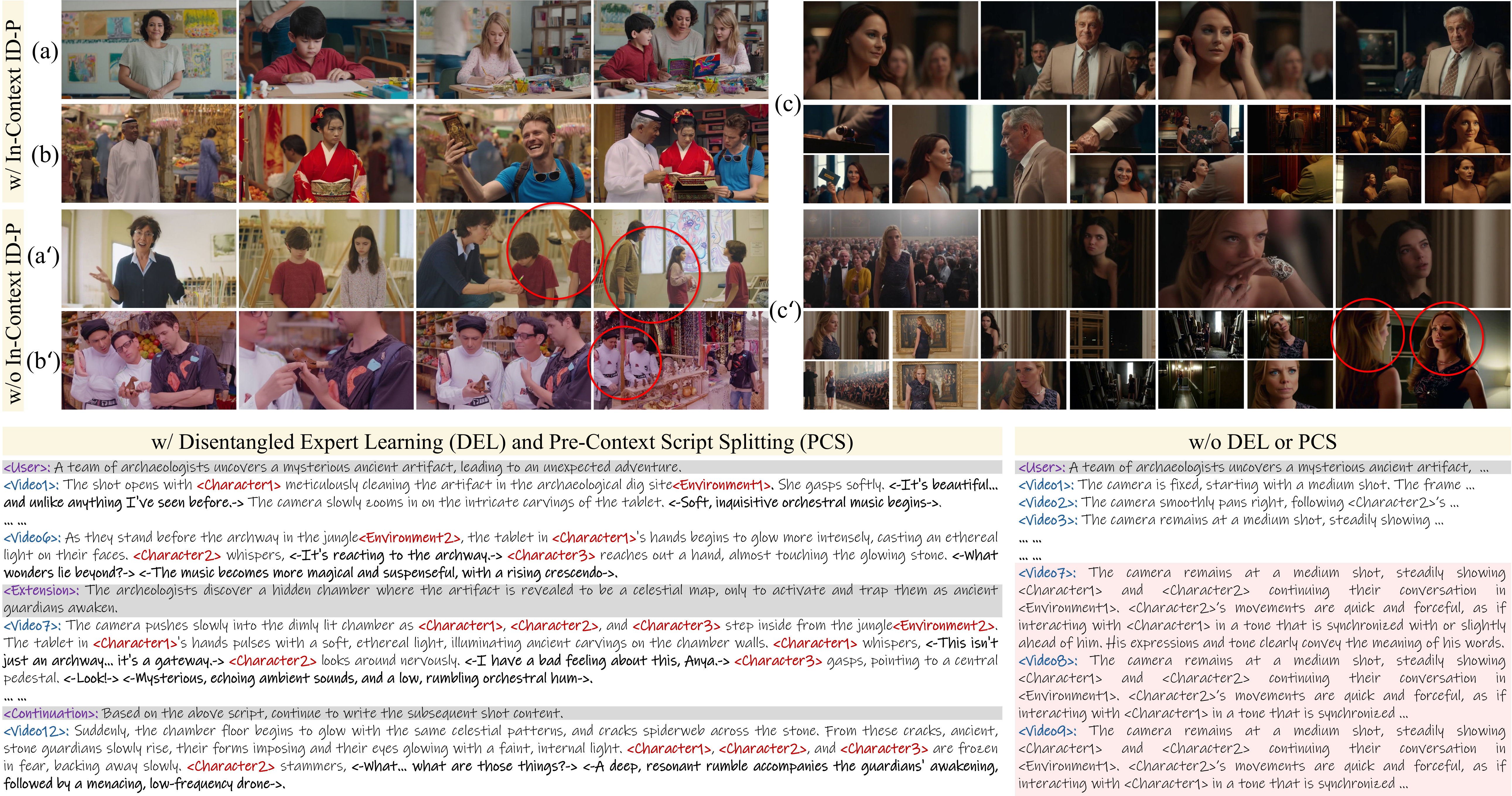}
\end{center}
   \caption{\textbf{Ablation experiments.} In-Context ID Prompting contributes to preserving consistent multi-character identities across multi-shot and long-form sequences; the red circles mark failure cases without it. Pre-Context Script Splitting ensures coherent and flexible script extension and continuation; the red box highlights content repetition issues when it is absent.}
\label{fig:exp_abl}
\end{figure}

\vspace{1mm}
\noindent\textbf{Long-form script generation.} Figure~\ref{fig:exp_sota2} shows the comparison on long-form script generation. StoryDiffusion maintains basic character consistency but suffers from limited scene variation, leading to repetitive visual patterns and copy-and-paste artifacts across shots. Story2Board and SEED-Story exhibit larger inconsistencies, and both models struggle to keep stable visual styles or character identities, producing fragmented transitions and weakening narrative coherence. In contrast, UniMAGE effectively models long-range temporal structure and follows the plot progression with much higher fidelity. With unified multimodal reasoning and stable ID conditioning, UniMAGE maintains consistent visual style and character appearance throughout the sequence, resulting in more coherent, diverse, and cinematic long-form storytelling.

\vspace{1mm}
\noindent\textbf{Multimodal script generation (vs. SEED-Story).} Although SEED-Story enables unified text–image script generation, its training is limited to only three animation datasets, resulting in poor generalization beyond stylized domains. As shown in Figure~\ref{fig:exp_seed}, its outputs often exhibit weak narrative logic and noticeably lower image quality. In contrast, UniMAGE generates more coherent plots with clearer causal structure, and produces significantly higher-quality keyframe images. Owing to its diverse training data and unified director architecture, UniMAGE generalizes well across different genres, visual styles, and storytelling formats.

\vspace{1mm}
\noindent\textbf{Ablation study for In-Context ID Prompting.} Figure~\ref{fig:exp_abl} presents the ablation results of the proposed In-Context ID Prompting strategy. With this mechanism, UniMAGE can reliably associate each visual entity with its corresponding textual identity, enabling stable character appearance across shots. When the strategy is removed, identity cues become ambiguous during scene transitions or long temporal spans, leading to mixed appearance and inconsistent depictions, as marked by the red circles. This confirms that explicit ID conditioning plays a crucial role in achieving robust multi-character consistency in long-form generation.

\vspace{1mm}
\noindent\textbf{Ablation study for Pre-Context Script Splitting.} As shown in the bottom part of Figure~\ref{fig:exp_abl}, the Pre-Context Script Splitting strategy significantly improves UniMAGE's ability to perform script extension and continuation. By training the understanding branch to generate narratives from partial context, the model learns to follow existing plot structures and introduce new developments naturally. In contrast, the interleaved-only training baseline frequently results in repetitive descriptions and weakened narrative flow, demonstrating the necessity of explicit pre-context conditioning for coherent long-form script generation.

\subsection{Quantitative Results}
Table~\ref{tab:1} reports the quantitative results on ViStoryBench. UniMAGE achieves the best overall performance, particularly in the consistency-related metrics. It obtains the highest CIDS (59.2) and OCCM (88.07), demonstrating strong character identity preservation across shots. UniMAGE also achieves a large improvement in Alignment (80.8), outperforming previous methods by a clear margin, indicating substantially better adherence to narrative prompts. While some baselines show competitive scores in isolated metrics (e.g., SEED-Story in CSD or StoryDiffusion in image quality and aesthetics), none offer balanced performance across consistency, narrative alignment, and visual quality as UniMAGE does. Additionally, the notably high CSD score of SEED-Story is mainly due to its overfitting to a narrow set of animation-style datasets, which leads to strong stylistic consistency but poor generalization across broader visual domains.

The comparison between “UniMAGE w/o ID-P” and the full UniMAGE confirms the importance of the In-Context ID Prompting strategy. Removing ID-P significantly degrades CSD, CIDS, and OCCM, indicating weaker identity stability and scene coherence. This demonstrates that explicit ID conditioning is crucial for maintaining consistent multi-character representation in long-form storytelling.

\begin{table}[t]
\small{
\caption{ \textbf{Quantitative comparisons and ablative experiments.} The best results are bolded, and the second-best are underlined. UniMAGE achieves superior performance in consistency-related metrics and alignment.}%
\label{tab:1}
\begin{tabular}{ l | c c c c c c } 
\toprule[1.5pt]
		\textbf{Methods} & \textbf{CSD$^{self}$}$_{\uparrow}$ & \textbf{CIDS$^{self}$}$_{\uparrow}$ & \textbf{Alignment}$_{\uparrow}$ & \textbf{OCCM}$_{\uparrow}$ & \textbf{Inception}$_{\uparrow}$ & \textbf{Aesthetics}$_{\uparrow}$  \\
            \midrule[0.5pt]
            TheaterGen~\citep{cheng2024theatergen} & 40.4 & 53.3 & 37.9 & 84.4 & 14.88 & 4.90 \\
            Story-Adapter & \underline{73.7} & 56.4 & 58.9 & 84.9 & 13.73 & 4.89 \\
            Story2Board~\citep{dinkevich2025story2board} & 55.2 & 51.4 & 56.7 & 85.5 & \underline{15.26} & \underline{5.11} \\
            StoryDiffusion~\citep{zhou2024storydiffusion} & 63.5 & \underline{57.0} & 59.7 & 85.3 & \textbf{15.71} & \textbf{5.76} \\
            SEED-Story~\citep{yang2025seed} & \textbf{74.9} & 48.7 & 29.5 & 85.9 & 6.33 & 3.84 \\
            \midrule[0.5pt]
            UniMAGE w/o ID-P  & 52.6 & 55.6 & \underline{62.5} & \underline{87.00} & 12.06 & 4.44  \\
            UniMAGE (Ours) & 59.0 & \textbf{59.2} & \textbf{80.8} & \textbf{88.07} & 12.97 & 4.55  \\
		\bottomrule[1.5pt]
\end{tabular}}
\end{table}

\begin{figure}[]
\begin{center}
   \includegraphics[width=0.9\linewidth]{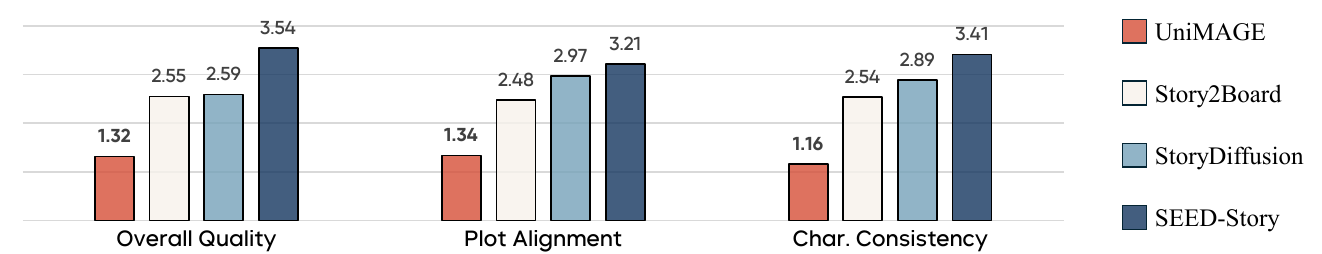}
\end{center}
   \caption{\textbf{User study.} Comparison of average rankings, where lower values correspond to higher quality.}
\label{fig:exp_user}
\end{figure}

\vspace{1mm}
\noindent\textbf{User study.} We conducted a user study with 50 volunteers, who evaluated 40 generated scripts by ranking UniMAGE and three baseline methods. For each comparison set, participants assessed the outputs in terms of overall quality, plot alignment, and character consistency, with an additional criterion of narrative logic specifically when comparing against SEED-Story. After removing invalid responses, the aggregated rankings are summarized in Figure~\ref{fig:exp_user}. UniMAGE achieves the highest preference across all criteria, including a GSB score of 0.72 for narrative logic, indicating a clear advantage in long-form story coherence. These results confirm that most participants prefer scripts generated by UniMAGE over those from existing baselines.
\section{Conclusions}

In this work, we introduced UniMAGE, a unified director model that integrates the traditionally disjoint processes of script drafting and keyframe generation into a single, coherent framework. By leveraging the Mixture-of-Transformers architecture, UniMAGE bridges the gap between textual reasoning and visual imagination, thereby enabling users to produce long-context, multi-shot narratives with logical and visual coherence. Central to our approach are two synergistic training paradigms: Interleaved Concept Learning, which fosters a joint understanding of narrative concepts through text–image interleaving, and Disentangled Expert Learning, which decouples script and keyframe generation to enhance both creativity and structural consistency. Further supported by In-Context ID Prompting and Pre-Context Script Splitting, UniMAGE demonstrates strong capability in maintaining character identity, storyline continuity, and visual alignment across extended sequences. Experimental evaluations confirm that UniMAGE achieves state-of-the-art results among open-source systems, establishing a foundation for the next generation of AI-driven film creation.

\vspace{1mm}
\noindent\textbf{Limitations.} UniMAGE is designed primarily to enhance narrative coherence and maintain strong visual consistency across long-form scripts. However, several higher-level dimensions of filmmaking—such as emotional pacing, stylistic cinematography, and fine-grained control over directorial intent—are not yet fully addressed. Extending UniMAGE toward richer cinematic understanding and more expressive narrative control remains an important direction for future work.

\vspace{1mm}
\noindent\textbf{Declaration.} This paper is conducted solely for research purposes, and the described technology has not been incorporated into any ByteDance products. All figures and scripts presented in this paper are generated by AI models. All human faces appearing in the images are also AI-generated.

\clearpage

\bibliographystyle{plainnat}
\bibliography{main}

@misc{veo3,
note = {\url{https://storage.googleapis.com/deepmind-media/veo/Veo-3-Tech-Report.pdf}},
title = {Veo: a text-to-video generation system},
author = {Google DeepMind},
year={2025}
}

@misc{sora2,
note = {\url{https://cdn.openai.com/pdf/50d5973c-c4ff-4c2d-986f-c72b5d0ff069/sora_2_system_card.pdf}},
title = {Sora 2 System Card},
author = {OpenAI},
year={2025}
}

@misc{klingai,
note = {\url{https://klingai.com}},
title = {Kling AI: Next-Generation AI Creative Studio},
author = {Kuaishou},
year={2025}
}

@misc{Gemini,
note = {\url{https://deepmind.google/models/gemini/pro/}},
title = {Gemini 2.5 Pro},
author = {Google DeepMind},
year={2025}
}

@article{huang2025filmaster,
  title={FilMaster: Bridging Cinematic Principles and Generative AI for Automated Film Generation},
  author={Huang, Kaiyi and Huang, Yukun and Wang, Xintao and Lin, Zinan and Ning, Xuefei and Wan, Pengfei and Zhang, Di and Wang, Yu and Liu, Xihui},
  journal={arXiv preprint arXiv:2506.18899},
  year={2025}
}

@article{wu2025automated,
  title={Automated movie generation via multi-agent cot planning},
  author={Wu, Weijia and Zhu, Zeyu and Shou, Mike Zheng},
  journal={arXiv preprint arXiv:2503.07314},
  year={2025}
}

@article{xiao2025captain,
  title={Captain cinema: Towards short movie generation},
  author={Xiao, Junfei and Yang, Ceyuan and Zhang, Lvmin and Cai, Shengqu and Zhao, Yang and Guo, Yuwei and Wetzstein, Gordon and Agrawala, Maneesh and Yuille, Alan and Jiang, Lu},
  journal={arXiv preprint arXiv:2507.18634},
  year={2025}
}

@inproceedings{long2024videostudio,
  title={Videostudio: Generating consistent-content and multi-scene videos},
  author={Long, Fuchen and Qiu, Zhaofan and Yao, Ting and Mei, Tao},
  booktitle={European Conference on Computer Vision},
  pages={468--485},
  year={2024},
  organization={Springer}
}

@article{deng2025emerging,
  title={Emerging properties in unified multimodal pretraining},
  author={Deng, Chaorui and Zhu, Deyao and Li, Kunchang and Gou, Chenhui and Li, Feng and Wang, Zeyu and Zhong, Shu and Yu, Weihao and Nie, Xiaonan and Song, Ziang and others},
  journal={arXiv preprint arXiv:2505.14683},
  year={2025}
}

@article{xie2024show,
  title={Show-o: One single transformer to unify multimodal understanding and generation},
  author={Xie, Jinheng and Mao, Weijia and Bai, Zechen and Zhang, David Junhao and Wang, Weihao and Lin, Kevin Qinghong and Gu, Yuchao and Chen, Zhijie and Yang, Zhenheng and Shou, Mike Zheng},
  journal={arXiv preprint arXiv:2408.12528},
  year={2024}
}

@article{zhou2024storydiffusion,
  title={Storydiffusion: Consistent self-attention for long-range image and video generation},
  author={Zhou, Yupeng and Zhou, Daquan and Cheng, Ming-Ming and Feng, Jiashi and Hou, Qibin},
  journal={Advances in Neural Information Processing Systems},
  volume={37},
  pages={110315--110340},
  year={2024}
}

@article{dinkevich2025story2board,
  title={Story2Board: A Training-Free Approach for Expressive Storyboard Generation},
  author={Dinkevich, David and Levy, Matan and Avrahami, Omri and Samuel, Dvir and Lischinski, Dani},
  journal={arXiv preprint arXiv:2508.09983},
  year={2025}
}

@inproceedings{li2024anim,
  title={Anim-director: A large multimodal model powered agent for controllable animation video generation},
  author={Li, Yunxin and Shi, Haoyuan and Hu, Baotian and Wang, Longyue and Zhu, Jiashun and Xu, Jinyi and Zhao, Zhen and Zhang, Min},
  booktitle={SIGGRAPH Asia 2024 Conference Papers},
  pages={1--11},
  year={2024}
}

@article{xu2025filmagent,
  title={Filmagent: A multi-agent framework for end-to-end film automation in virtual 3d spaces},
  author={Xu, Zhenran and Wang, Longyue and Wang, Jifang and Li, Zhouyi and Shi, Senbao and Yang, Xue and Wang, Yiyu and Hu, Baotian and Yu, Jun and Zhang, Min},
  journal={arXiv preprint arXiv:2501.12909},
  year={2025}
}

@inproceedings{ruan2023mm,
  title={Mm-diffusion: Learning multi-modal diffusion models for joint audio and video generation},
  author={Ruan, Ludan and Ma, Yiyang and Yang, Huan and He, Huiguo and Liu, Bei and Fu, Jianlong and Yuan, Nicholas Jing and Jin, Qin and Guo, Baining},
  booktitle={Proceedings of the IEEE/CVF Conference on Computer Vision and Pattern Recognition},
  pages={10219--10228},
  year={2023}
}

@article{liu2024syncflow,
  title={SyncFlow: Toward Temporally Aligned Joint Audio-Video Generation from Text},
  author={Liu, Haohe and Lan, Gael Le and Mei, Xinhao and Ni, Zhaoheng and Kumar, Anurag and Nagaraja, Varun and Wang, Wenwu and Plumbley, Mark D and Shi, Yangyang and Chandra, Vikas},
  journal={arXiv preprint arXiv:2412.15220},
  year={2024}
}

@article{zhao2025uniform,
  title={UniForm: A Unified Multi-Task Diffusion Transformer for Audio-Video Generation},
  author={Zhao, Lei and Feng, Linfeng and Ge, Dongxu and Chen, Rujin and Yi, Fangqiu and Zhang, Chi and Zhang, Xiao-Lei and Li, Xuelong},
  journal={arXiv preprint arXiv:2502.03897},
  year={2025}
}

@article{wang2025universe,
  title={UniVerse-1: Unified Audio-Video Generation via Stitching of Experts},
  author={Wang, Duomin and Zuo, Wei and Li, Aojie and Chen, Ling-Hao and Liao, Xinyao and Zhou, Deyu and Yin, Zixin and Dai, Xili and Jiang, Daxin and Yu, Gang},
  journal={arXiv preprint arXiv:2509.06155},
  year={2025}
}

@article{gao2025seedance,
  title={Seedance 1.0: Exploring the Boundaries of Video Generation Models},
  author={Gao, Yu and Guo, Haoyuan and Hoang, Tuyen and Huang, Weilin and Jiang, Lu and Kong, Fangyuan and Li, Huixia and Li, Jiashi and Li, Liang and Li, Xiaojie and others},
  journal={arXiv preprint arXiv:2506.09113},
  year={2025}
}

@article{wang2024emu3,
  title={Emu3: Next-token prediction is all you need},
  author={Wang, Xinlong and Zhang, Xiaosong and Luo, Zhengxiong and Sun, Quan and Cui, Yufeng and Wang, Jinsheng and Zhang, Fan and Wang, Yueze and Li, Zhen and Yu, Qiying and others},
  journal={arXiv preprint arXiv:2409.18869},
  year={2024}
}

@article{team2024chameleon,
  title={Chameleon: Mixed-modal early-fusion foundation models},
  author={Team, Chameleon},
  journal={arXiv preprint arXiv:2405.09818},
  year={2024}
}

@article{zhou2024transfusion,
  title={Transfusion: Predict the next token and diffuse images with one multi-modal model},
  author={Zhou, Chunting and Yu, Lili and Babu, Arun and Tirumala, Kushal and Yasunaga, Michihiro and Shamis, Leonid and Kahn, Jacob and Ma, Xuezhe and Zettlemoyer, Luke and Levy, Omer},
  journal={arXiv preprint arXiv:2408.11039},
  year={2024}
}

@article{minaee2024large,
  title={Large language models: A survey},
  author={Minaee, Shervin and Mikolov, Tomas and Nikzad, Narjes and Chenaghlu, Meysam and Socher, Richard and Amatriain, Xavier and Gao, Jianfeng},
  journal={arXiv preprint arXiv:2402.06196},
  year={2024}
}

@article{liu2022flow,
  title={Flow straight and fast: Learning to generate and transfer data with rectified flow},
  author={Liu, Xingchao and Gong, Chengyue and Liu, Qiang},
  journal={arXiv preprint arXiv:2209.03003},
  year={2022}
}

@inproceedings{esser2024scaling,
  title={Scaling rectified flow transformers for high-resolution image synthesis},
  author={Esser, Patrick and Kulal, Sumith and Blattmann, Andreas and Entezari, Rahim and M{\"u}ller, Jonas and Saini, Harry and Levi, Yam and Lorenz, Dominik and Sauer, Axel and Boesel, Frederic and others},
  booktitle={Forty-first international conference on machine learning},
  year={2024}
}

@article{wei2022chain,
  title={Chain-of-thought prompting elicits reasoning in large language models},
  author={Wei, Jason and Wang, Xuezhi and Schuurmans, Dale and Bosma, Maarten and Xia, Fei and Chi, Ed and Le, Quoc V and Zhou, Denny and others},
  journal={Advances in neural information processing systems},
  volume={35},
  pages={24824--24837},
  year={2022}
}

@article{fang2025got,
  title={Got: Unleashing reasoning capability of multimodal large language model for visual generation and editing},
  author={Fang, Rongyao and Duan, Chengqi and Wang, Kun and Huang, Linjiang and Li, Hao and Yan, Shilin and Tian, Hao and Zeng, Xingyu and Zhao, Rui and Dai, Jifeng and others},
  journal={arXiv preprint arXiv:2503.10639},
  year={2025}
}

@article{hui2024qwen2,
  title={Qwen2. 5-coder technical report},
  author={Hui, Binyuan and Yang, Jian and Cui, Zeyu and Yang, Jiaxi and Liu, Dayiheng and Zhang, Lei and Liu, Tianyu and Zhang, Jiajun and Yu, Bowen and Lu, Keming and others},
  journal={arXiv preprint arXiv:2409.12186},
  year={2024}
}

@article{cheng2024theatergen,
  title={Theatergen: Character management with llm for consistent multi-turn image generation},
  author={Cheng, Junhao and Yin, Baiqiao and Cai, Kaixin and Huang, Minbin and Li, Hanhui and He, Yuxin and Lu, Xi and Li, Yue and Li, Yifei and Cheng, Yuhao and others},
  journal={arXiv preprint arXiv:2404.18919},
  year={2024}
}

@article{shi2025animaker,
  title={AniMaker: Automated Multi-Agent Animated Storytelling with MCTS-Driven Clip Generation},
  author={Shi, Haoyuan and Li, Yunxin and Chen, Xinyu and Wang, Longyue and Hu, Baotian and Zhang, Min},
  journal={arXiv preprint arXiv:2506.10540},
  year={2025}
}

@inproceedings{yang2025seed,
  title={Seed-story: Multimodal long story generation with large language model},
  author={Yang, Shuai and Ge, Yuying and Li, Yang and Chen, Yukang and Ge, Yixiao and Shan, Ying and Chen, Ying-Cong},
  booktitle={Proceedings of the IEEE/CVF International Conference on Computer Vision},
  pages={1850--1860},
  year={2025}
}

@misc{mao2024story_adapter,
  title={{Story-Adapter: A Training-free Iterative Framework for Long Story Visualization}},
  author={Mao, Jiawei and Huang, Xiaoke and Xie, Yunfei and Chang, Yuanqi and Hui, Mude and Xu, Bingjie and Zhou, Yuyin},
  journal={arXiv},
  volume={abs/2410.06244},
  year={2024},
}

@article{zhuang2025vistorybench,
  title={Vistorybench: Comprehensive benchmark suite for story visualization},
  author={Zhuang, Cailin and Huang, Ailin and Cheng, Wei and Wu, Jingwei and Hu, Yaoqi and Liao, Jiaqi and Wang, Hongyuan and Liao, Xinyao and Cai, Weiwei and Xu, Hengyuan and others},
  journal={arXiv preprint arXiv:2505.24862},
  year={2025}
}

@article{wu2023qalign,
  title={Q-Align: Teaching LMMs for Visual Scoring via Discrete Text-Defined Levels},
  author={Wu, Haoning and Zhang, Zicheng and Zhang, Weixia and Chen, Chaofeng and Li, Chunyi and Liao, Liang and Wang, Annan and Zhang, Erli and Sun, Wenxiu and Yan, Qiong and Min, Xiongkuo and Zhai, Guangtai and Lin, Weisi},
  journal={arXiv preprint arXiv:2312.17090},
  year={2023},
  institution={Nanyang Technological University and Shanghai Jiao Tong University and Sensetime Research},
  note={Equal Contribution by Wu, Haoning and Zhang, Zicheng. Project Lead by Wu, Haoning. Corresponding Authors: Zhai, Guangtai and Lin, Weisi.}
}

@inproceedings{dehghani2023scaling,
  title={Scaling vision transformers to 22 billion parameters},
  author={Dehghani, Mostafa and Djolonga, Josip and Mustafa, Basil and Padlewski, Piotr and Heek, Jonathan and Gilmer, Justin and Steiner, Andreas Peter and Caron, Mathilde and Geirhos, Robert and Alabdulmohsin, Ibrahim and others},
  booktitle={International conference on machine learning},
  pages={7480--7512},
  year={2023},
  organization={PMLR}
}

@article{tschannen2025siglip,
  title={Siglip 2: Multilingual vision-language encoders with improved semantic understanding, localization, and dense features},
  author={Tschannen, Michael and Gritsenko, Alexey and Wang, Xiao and Naeem, Muhammad Ferjad and Alabdulmohsin, Ibrahim and Parthasarathy, Nikhil and Evans, Talfan and Beyer, Lucas and Xia, Ye and Mustafa, Basil and others},
  journal={arXiv preprint arXiv:2502.14786},
  year={2025}
}

@article{dehghani2023patch,
  title={Patch n’pack: Navit, a vision transformer for any aspect ratio and resolution},
  author={Dehghani, Mostafa and Mustafa, Basil and Djolonga, Josip and Heek, Jonathan and Minderer, Matthias and Caron, Mathilde and Steiner, Andreas and Puigcerver, Joan and Geirhos, Robert and Alabdulmohsin, Ibrahim M and others},
  journal={Advances in Neural Information Processing Systems},
  volume={36},
  pages={2252--2274},
  year={2023}
}

@misc{flux,
note = {\url{https://github.com/black-forest-labs/flux}},
title = {Flux},
author = {Black Forest Labs},
year={2024}
}

\clearpage

\beginappendix
\renewcommand\thesection{\Alph{section}}
\setcounter{section}{0}


\section{Mixture-of-Transformers Architecture}
\label{sec:sup_mot}

\begin{figure}[h]
\setlength{\abovecaptionskip}{-.2cm}
\begin{center}
   \includegraphics[width=1.0\linewidth]{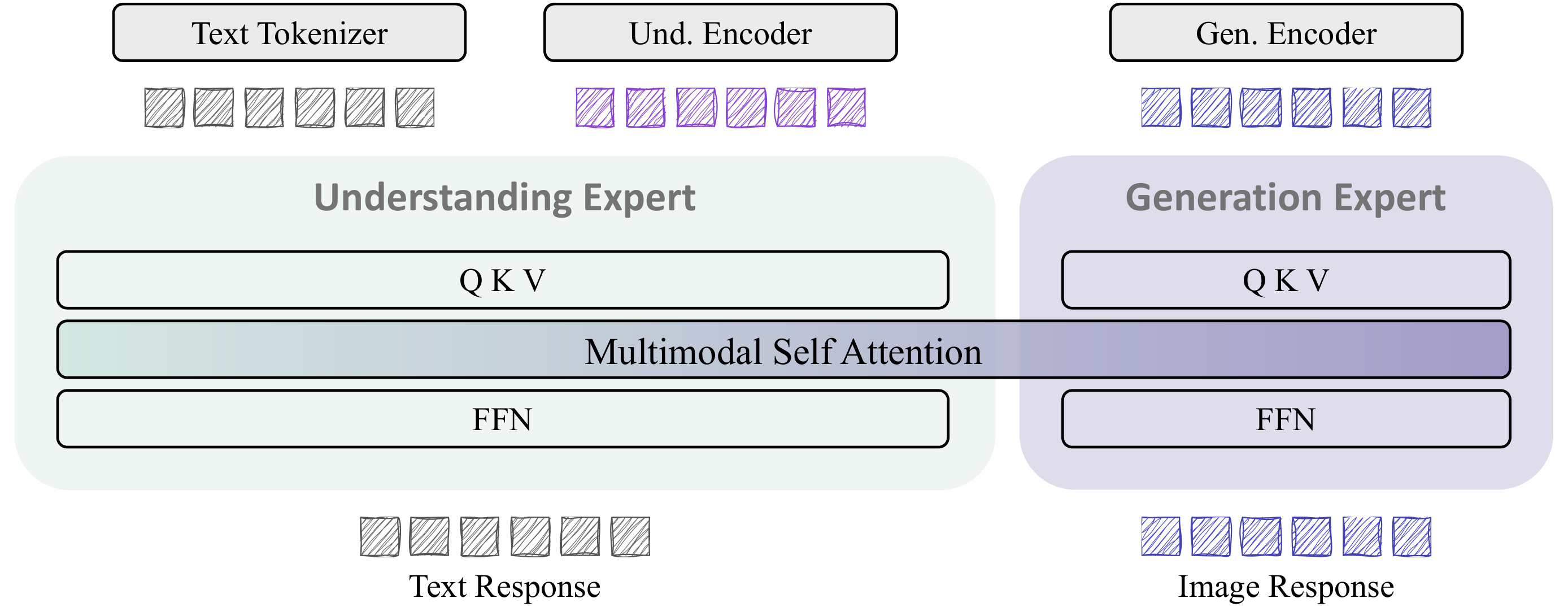}
\end{center}
   \caption{\textbf{Architecture of Mixture-of-Transformers.} MoT employs two Transformer experts to process understanding and generation information, and all tokens are processed through a shared multi-modal self-attention in each Transformer block. Two distinct encoders, i.e., ViT and VAE, are adopted separately to capture semantic content and low-level information for image understanding and generation tasks.}
\label{fig:sup_mot}
\end{figure}

Our UniMAGE framework is based on the Mixture-of-Transformers (MoT) architecture introduced in Bagel~\citep{deng2025emerging}, as shown in Figure \ref{fig:sup_mot}. MoT combines two transformers through shared multimodal self-attention layers. The transformer parameters are initialized from the Qwen 2.5 LLM~\citep{hui2024qwen2}. To improve training stability, each attention block includes the QK-Norm~\citep{dehghani2023scaling}.

The visual information is represented from two aspects:

(1) Visual understanding — handled by a ViT encoder. It uses SigLIP2-so400m/14~\citep{tschannen2025siglip} with a fixed input resolution of 384 for initialization. The position embeddings are interpolated, and the maximum input size is set to $980 \times 980$. NaViT~\citep{dehghani2023patch} is incorporated to handle images in their native aspect ratios. A two-layer MLP connector aligns the ViT token dimensions with the LLM hidden states.

(2) Visual generation — managed by a pre-trained VAE model from FLUX~\citep{flux}, which maps images between pixel and latent spaces. The latent representation has a downsampling ratio of 8, and the latent channels is 16. A $2 \times 2$ patch embedding layer is used to further reduce spatial dimensions and align with the LLM hidden size.

Within MoT, text, ViT, and VAE tokens are interleaved according to the multimodal input structure. For tokens from the same sample, a generalized causal attention scheme is used. Tokens are divided into sequential splits by modality (text, ViT, or VAE). Each split can attend to all earlier splits, with causal attention applied to text tokens and bidirectional attention for vision tokens.

\section{Supplementary on Script Dataset}
\label{sec:sup_script}

The construction of a text–image interleaved script dataset presents significant challenges. Therefore, we detail our method for building such a dataset from multi-shot videos. Specifically, we curate a film dataset comprising approximately 800k multi-shot video clips, employing a video filtering strategy consistent with that adopted in Seedance 1.0~\citep{gao2025seedance}. We then utilize Gemini 2.5 Pro~\citep{Gemini} to generate captions for these video clips using the designed prompt. Subsequently, Q-Align~\citep{wu2023qalign} is employed to assess the image quality and aesthetics of keyframes, from which we select one keyframe image within the first 10\% of frames of each video clip. After filtering based on these quality metrics, approximately 450k multi-shot video clips are retained to construct the final text–image interleaved script dataset. To automatically generate user prompts, we define four types of prompt styles: (1) simple narrative, (2) abstract concept, (3) phrase splicing, and (4) spoken expression. The detailed prompt design is presented as follows.

\begin{figure}[h]
\setlength{\abovecaptionskip}{-.2cm}
\begin{center}
   \includegraphics[width=1.0\linewidth]{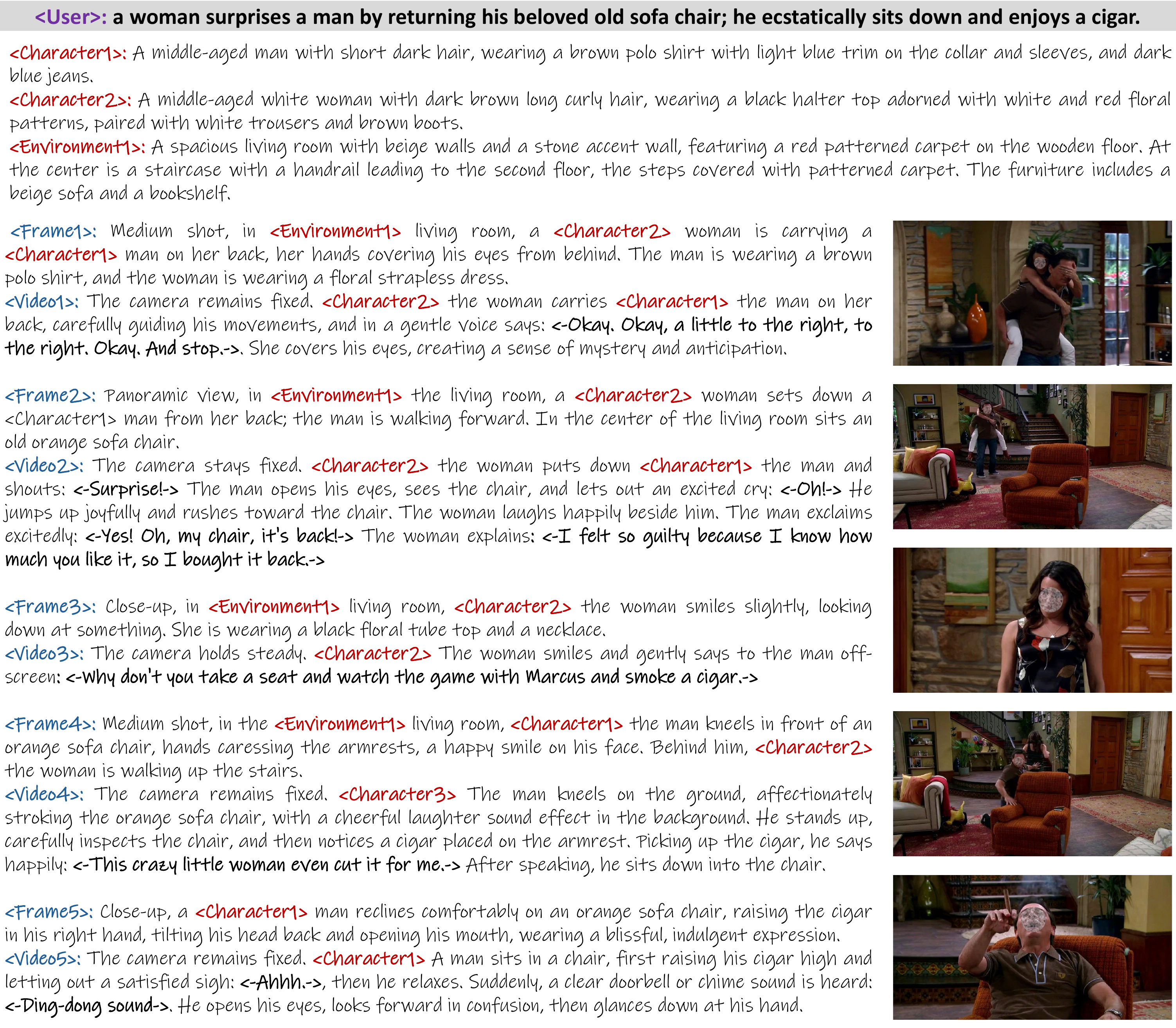}
\end{center}
   \caption{\textbf{Example of the text–image interleaved script dataset.} The dataset is constructed by captioning multi-shot video clips and extracting corresponding keyframes.}
\label{fig:sup_data}
\end{figure}

Figure \ref{fig:sup_data} illustrates an example of the text–image interleaved script dataset. For a detailed description of the script structure, please refer to \textit{Section 3.2 Script Structure} in the main paper.

\begin{center}
\begin{promptbox}[Multi-Shot Video Annotation Prompt]
\textbf{\# Role}\\
You are a professional video content analyst.

\vspace{6pt}
\textbf{\# Task} \\[3pt]
Your task is to accurately analyze the provided video file (including video and audio content) and provide structured annotation information according to my requirements. You need to understand both visual and auditory information and correlate them.\\[3pt]
Please annotate this video clip containing \texttt{\{NUM\_VIDEOS\}} sorted shots. The annotation format needs to be JSON, with the key-value pairs as follows:

\begin{lstlisting}
{{
"global_caption":
{{
    "first_frame_description": "Detailed visual element annotation of the first frame of the video, including a description of the camera angle, visual style, main characters, environment, and main objects, as well as the position and posture of each element. At most 100 words."
    "video_audio_description": "Detailed annotation of the entire audio and video, including a description of the video's camera movement, visual style, how the N objects interact in the M environments, including the manner, style, intensity, and specific actions of the interactions, conversation content, etc. At most 150 words."
    "key_character_description": "Detailed appearance description of the N objects (N people or items) and M environments contained in the entire video. The annotation results need to be indexed by serial numbers for the identified N objects and M environments. At the same time, please select the most representative frame for each object's appearance  to be used for subsequent extraction of the object's visual features. The overall format is, for example, {{Character1: caption:..., short_caption: a concise version of the caption, each object description does not exceed 10 words (E.g., middle-aged white male, short gray curly hair, black-framed glasses, wrinkles around the eyes/forehead, dark green button-down shirt), frame_id: {{shot: i, second: x}}, Character2: caption:..., short_caption:..., frame_id: {{shot: j, second: y}}, ...}}
}},
"shot_caption": "Covers the first frame appearance description of the current shot segment, and the video and audio description of the current shot segment. The ID indexes of the characters and environments that appear in all shots need to be aligned and consistent with the global annotations, appearing in the text annotation results involving characters in the format of \<Character1\>, \<Character2\>..."
{{
    [
    {{
        "first_frame_description": "A detailed description of the first frame of the current shot, including a description of the camera angle, visual style, main characters, environment, and main objects, as well as the position and posture of each element. At most 100 words."
        "video_audio_description": "A detailed description of the video and audio of the shot segment, at most 150 words. Please describe in order the video's camera movement, visual style, the appearance and specific events of the characters that appear, the appearance of the environment they are in, the interaction process between characters, environments, and objects, what the characters said and in what tone (keep the original spoken content, do not translate, and mark the spoken content with <- and ->), background sound description (style, content, and intensity, marked with <- and ->), and possible voice-overs, etc."
        "have_background_audio": Select from ["None", "Environment", "Music"], which respectively represent: no background sound, background environment sound, and background music sound. Except for the None tag, other tags can be selected in combination.
    }},
    ...
    ]
}},
"user_prompt": Use a sentence of no more than 40 words to describe global_caption and shot_caption. Please randomly select one of the following 4 styles to imitate and describe. This will be used as the input for the subsequent training of the video text annotation model. The output is global_caption and shot_caption. If shot_caption contains multiple shots, randomly choose whether to label each shot individually, for example, shot 1:..., shot 2:...
"user_prompt_style": The selected style number.
"user_prompt_has_shot": Whether to choose to caption each shot individually, true or false.
}}
\end{lstlisting}

The video and audio should be described simultaneously in chronological order.  
Do not describe the video first and then the audio.  
Only return the annotation results, do not return other content.  

\vspace{6pt}
The annotation results should all be in \texttt{\{LANGUAGE\}}, but please keep the spoken content in the original language.  
Only annotate the original content of the video, without inference or imagination.

\vspace{6pt}
User prompt styles to be selected:

1. Simple narrative, such as a cute anthropomorphic kitten, wearing an apron, holding a frying pan in one hand, frying braised pork in a pot, the frying sound is sizzling, the pot is smoking, and the smoke is lingering above the pot.

2. Abstract concepts, such as the ancient Qin Shihuang unified the six kingdoms.

3. Phrase splicing, such as a photo, a super large and expensive panda, light champagne color, close-up of facial features, wearing a suit, dancing in the square, medium shot, parallel perspective, slowly close up, bright colors, cheerful, extremely delicate pictures, ultra-high resolution, and movie effects.

4. Spoken expression, such as generating a video of a panda mother riding a shared bicycle and taking a panda baby to school, the panda needs to be realistic, not cartoon, and the background is a vegetable market.
\end{promptbox}
\end{center}

\section{Supplementary on Experiments}
\label{sec:sup_exp}
Due to the page limitations of the main text, it is not possible to display the complete generated multimodal scripts. The full versions of all results presented in the main paper are provided below. In addition, we include additional generated scripts to further demonstrate the generalizability of UniMAGE.

\section{Supplementary on Audio-Video Generation}
\label{sec:sup_av}
Returning to our original motivation, beyond story visualization, we aim to develop a unified director model capable of serving audio–video generation systems. Accordingly, we provide two video examples in the supplementary materials to demonstrate the capability of UniMAGE in facilitating automated audio and video generation. The video files are named ``UniMAGE\_Demo\_1.mp4'' and ``UniMAGE\_Demo\_2.mp4''. We sincerely encourage readers to view the paper in conjunction with these videos. These two video demonstrations are generated using the Veo3 I2V model~\citep{veo3}, with the scripts, character lines, and sound effect descriptions provided by UniMAGE. It is worth noting that, due to the current limitations of audio-video generation models, the timbre and facial features of characters may vary during the generation of long multi-camera videos. These issues warrant further investigation and improvement in future work.


\begin{figure}[t]
\begin{center}
   \includegraphics[width=1.0\linewidth]{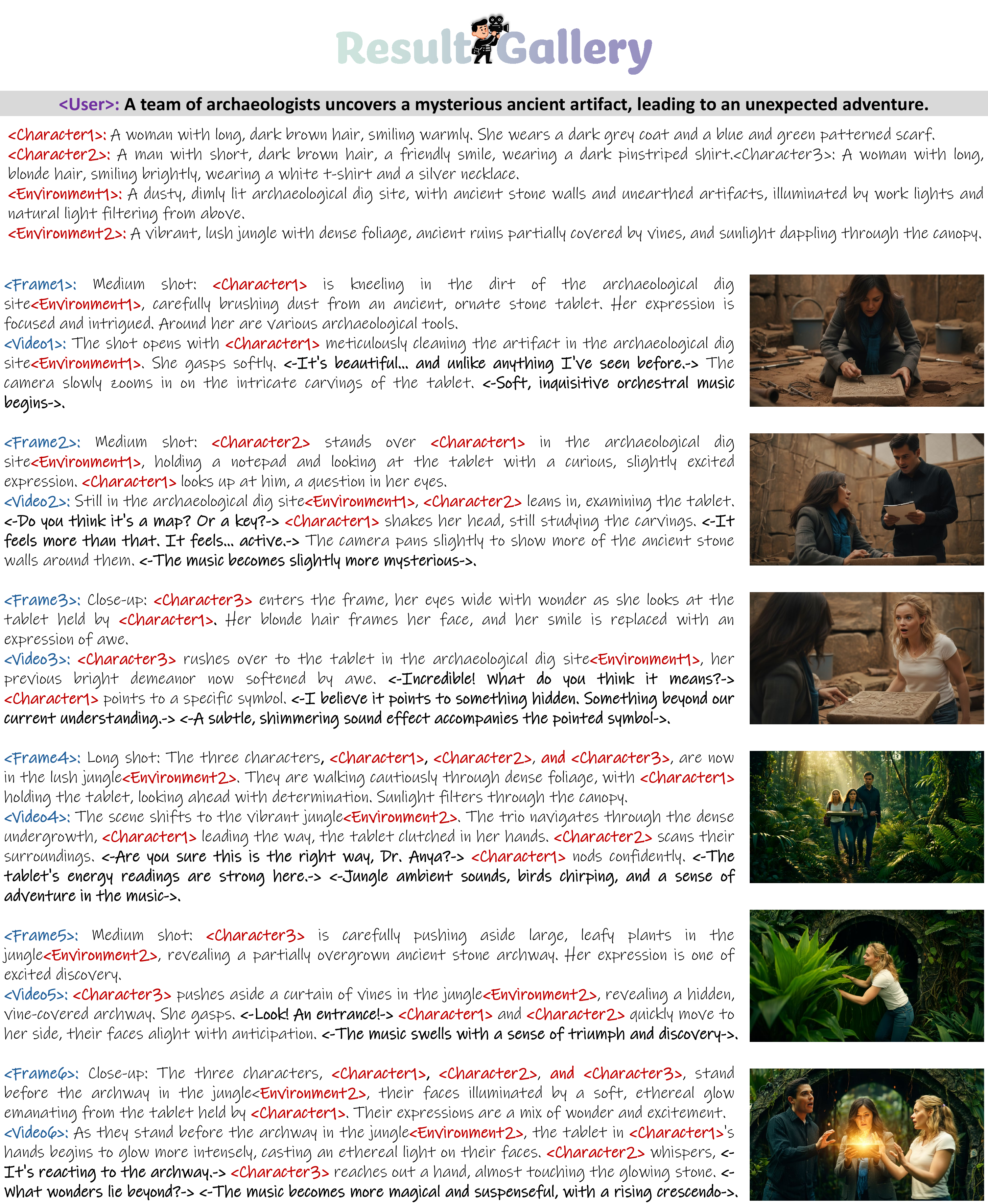}
\end{center}
\label{fig:sup_exp1}
\end{figure}

\begin{figure}[t]
\begin{center}
   \includegraphics[width=1.0\linewidth]{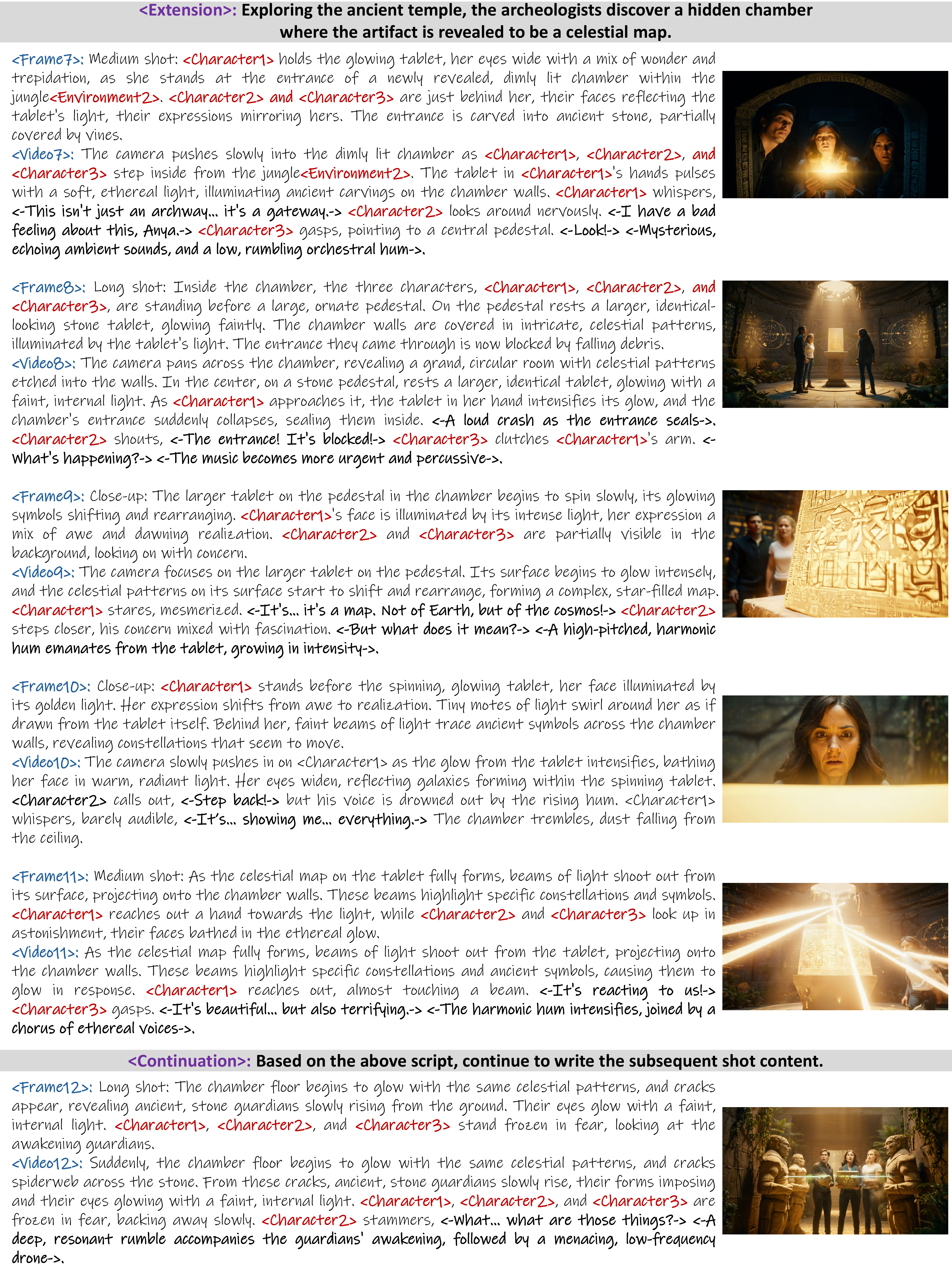}
\end{center}
\label{fig:sup_exp2}
\end{figure}

\begin{figure}[t]
\begin{center}
   \includegraphics[width=1.0\linewidth]{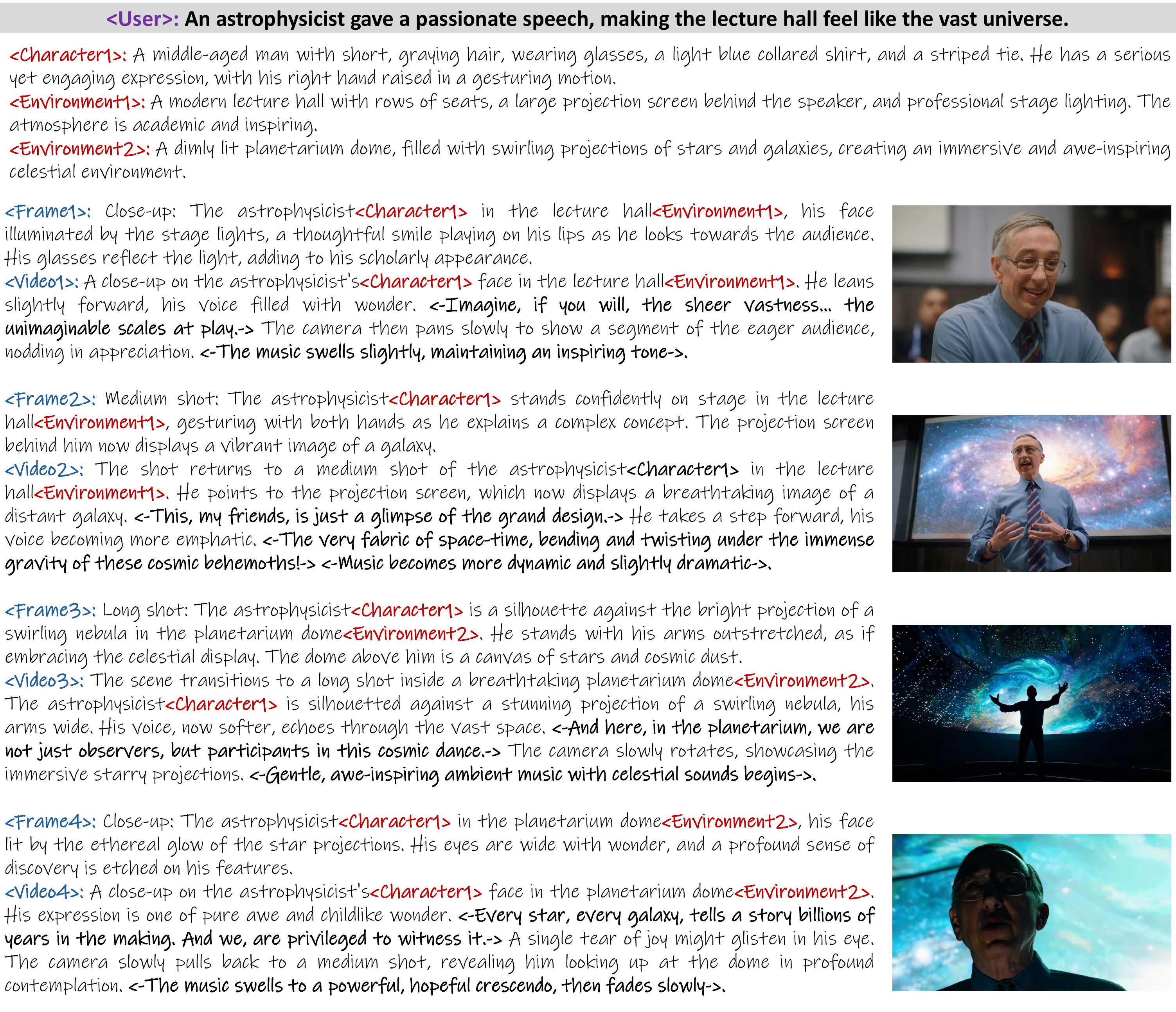}
\end{center}
\label{fig:sup_exp3}
\end{figure}

\begin{figure}[t]
\begin{center}
   \includegraphics[width=1.0\linewidth]{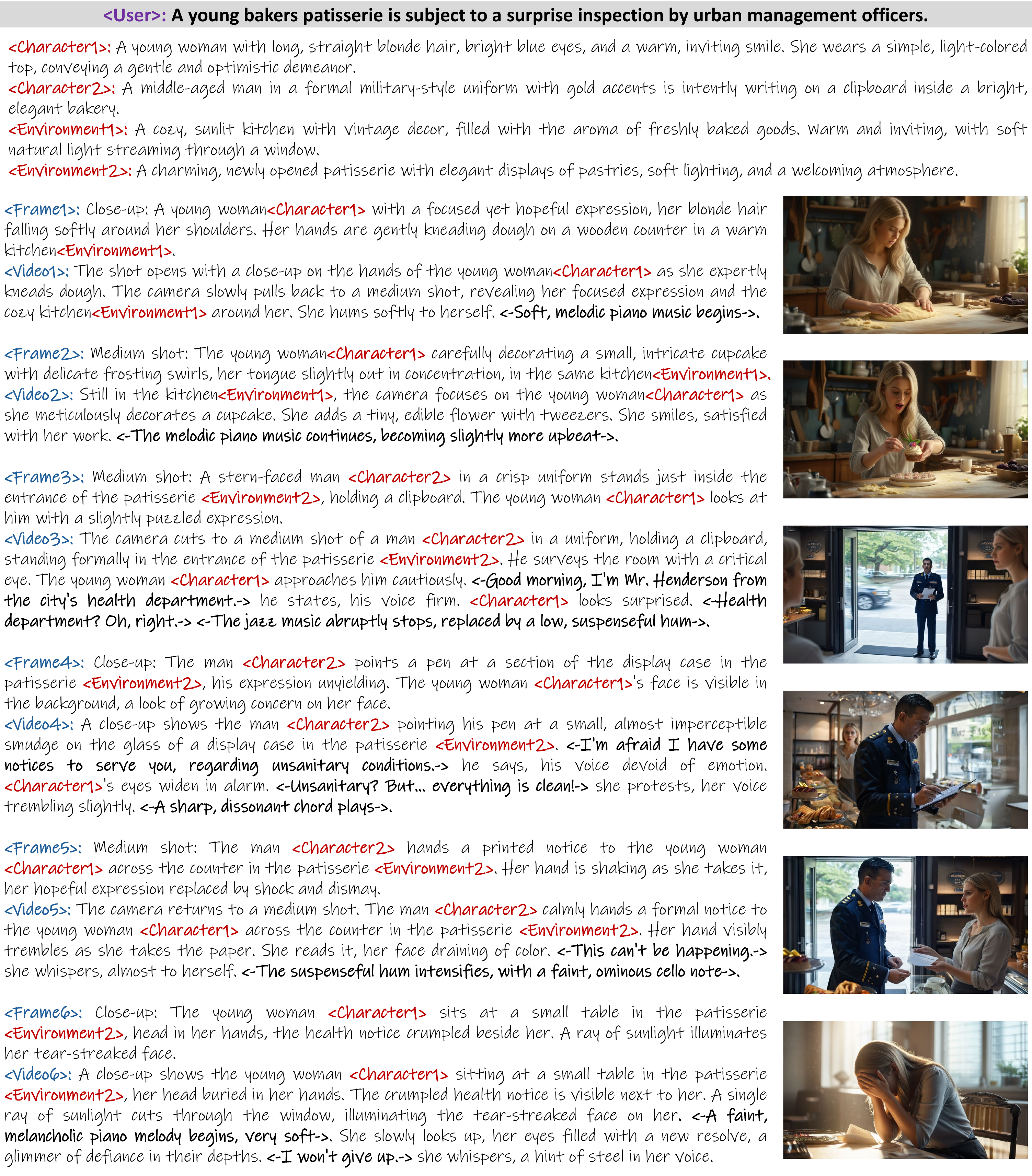}
\end{center}
\label{fig:sup_exp4}
\end{figure}

\begin{figure}[t]
\begin{center}
   \includegraphics[width=1.0\linewidth]{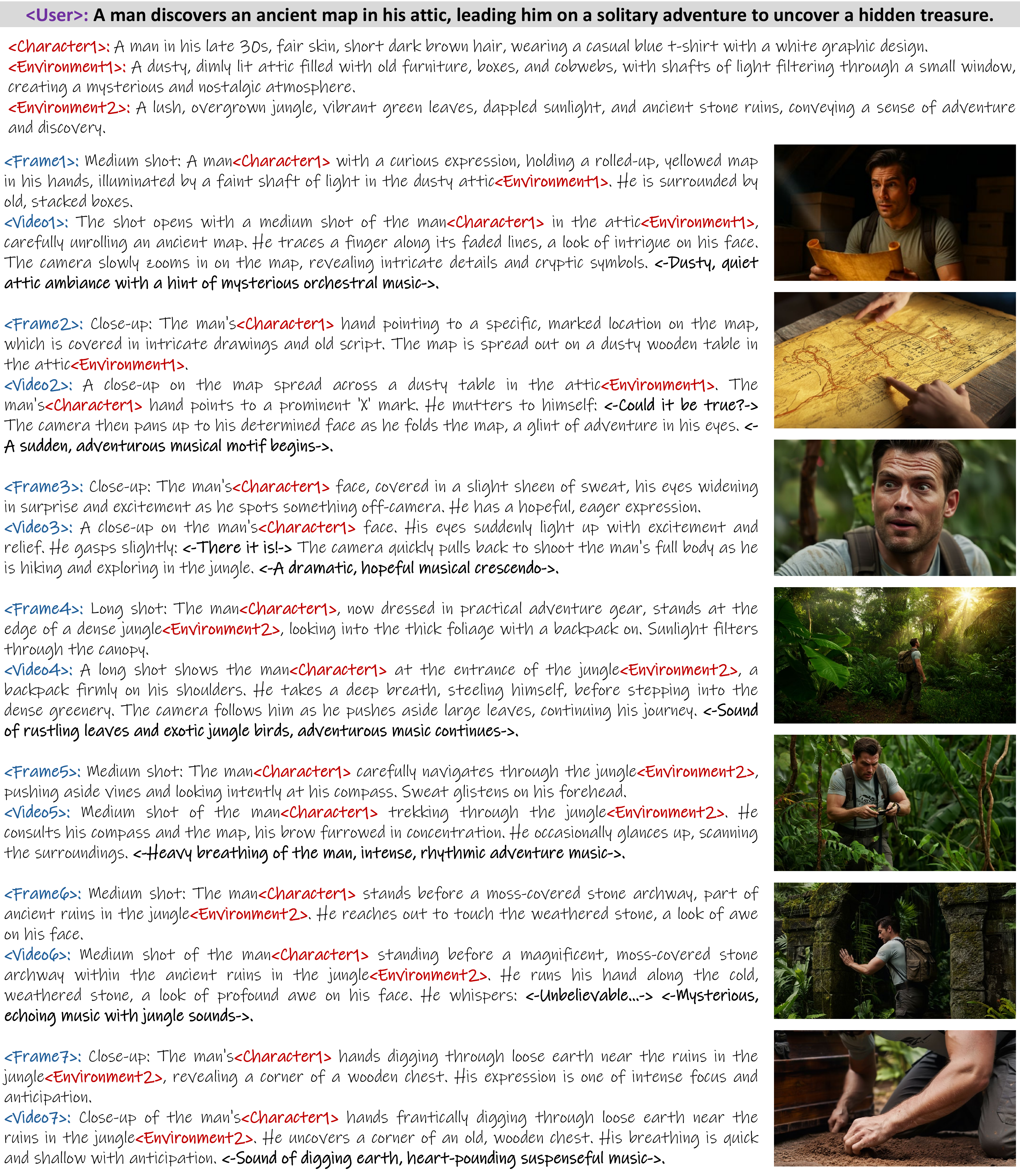}
\end{center}
\label{fig:sup_exp5}
\end{figure}

\begin{figure}[t]
\begin{center}
   \includegraphics[width=1.0\linewidth]{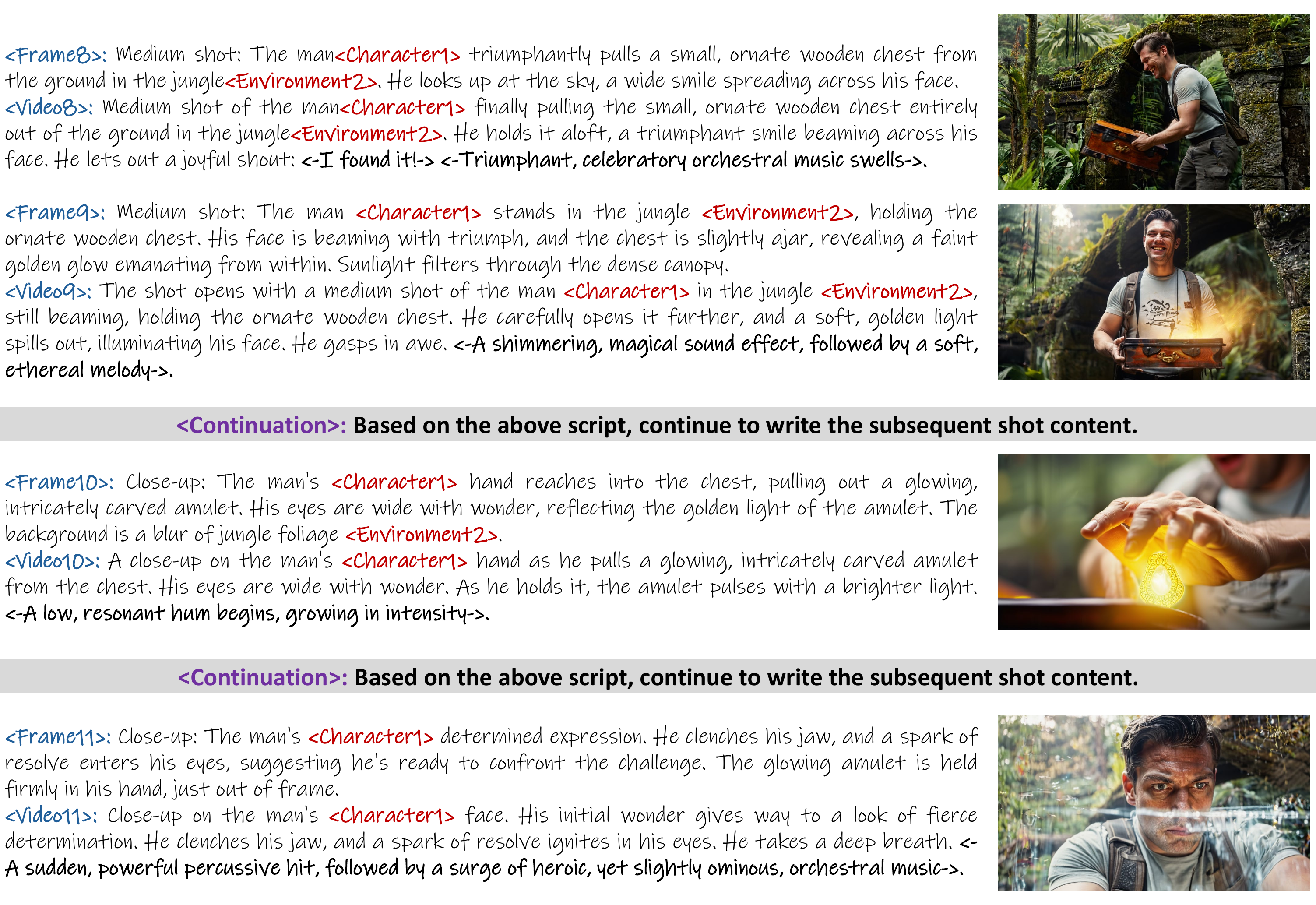}
\end{center}
\label{fig:sup_exp6}
\end{figure}

\begin{figure}[t]
\begin{center}
   \includegraphics[width=1.0\linewidth]{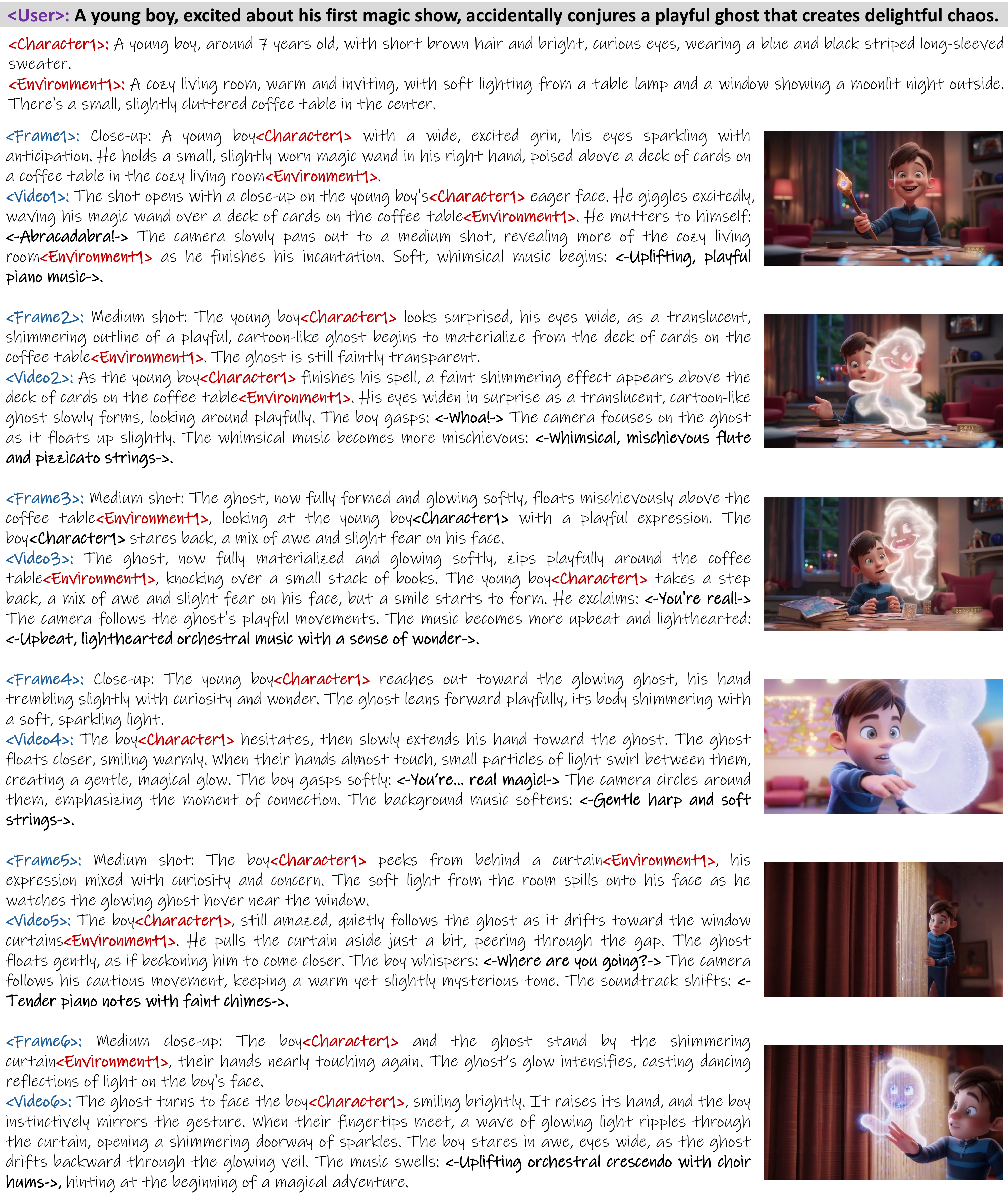}
\end{center}
\label{fig:sup_exp12}
\end{figure}

\begin{figure}[t]
\begin{center}
   \includegraphics[width=1.0\linewidth]{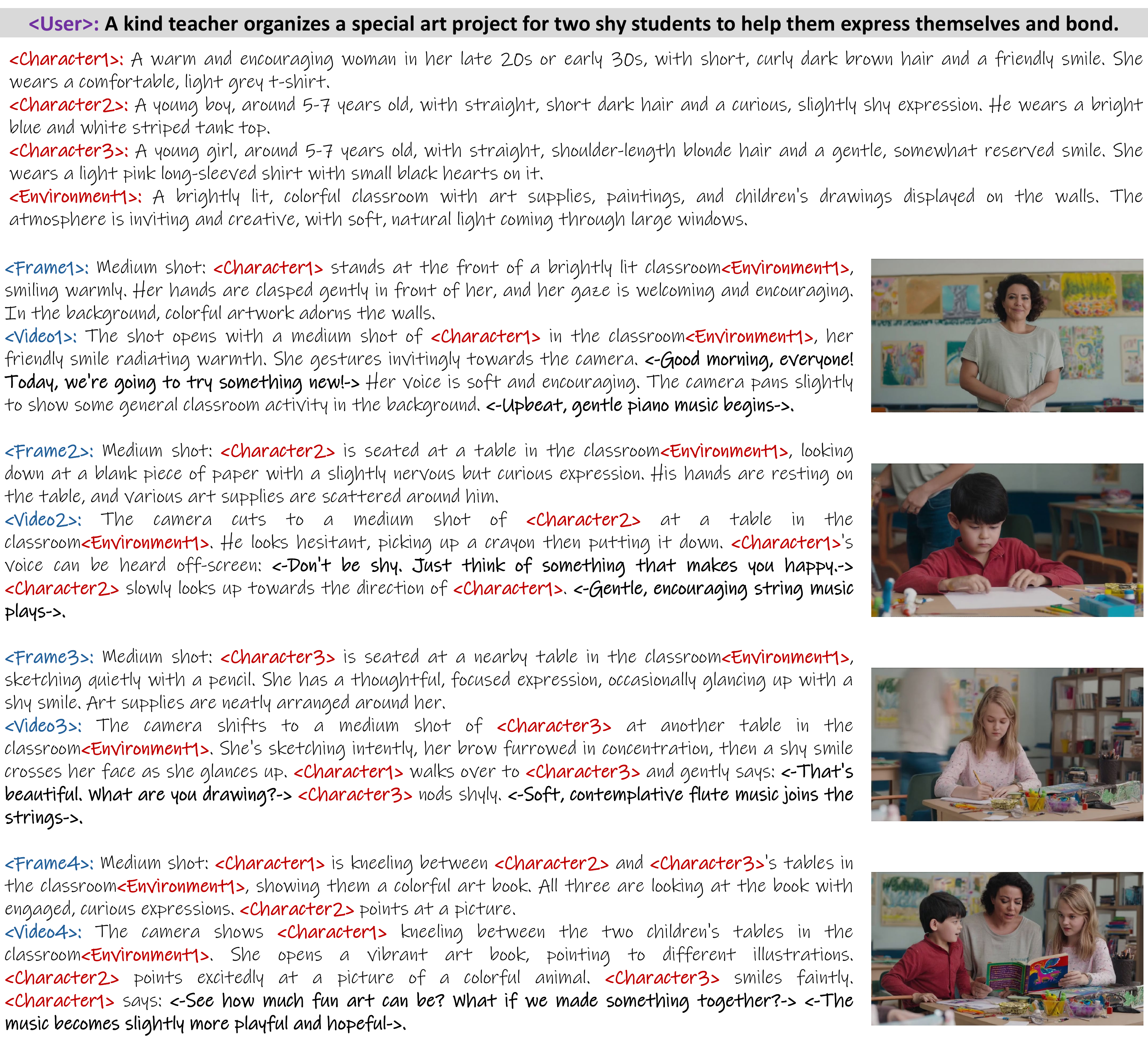}
\end{center}
\label{fig:sup_exp7}
\end{figure}

\begin{figure}[t]
\begin{center}
   \includegraphics[width=1.0\linewidth]{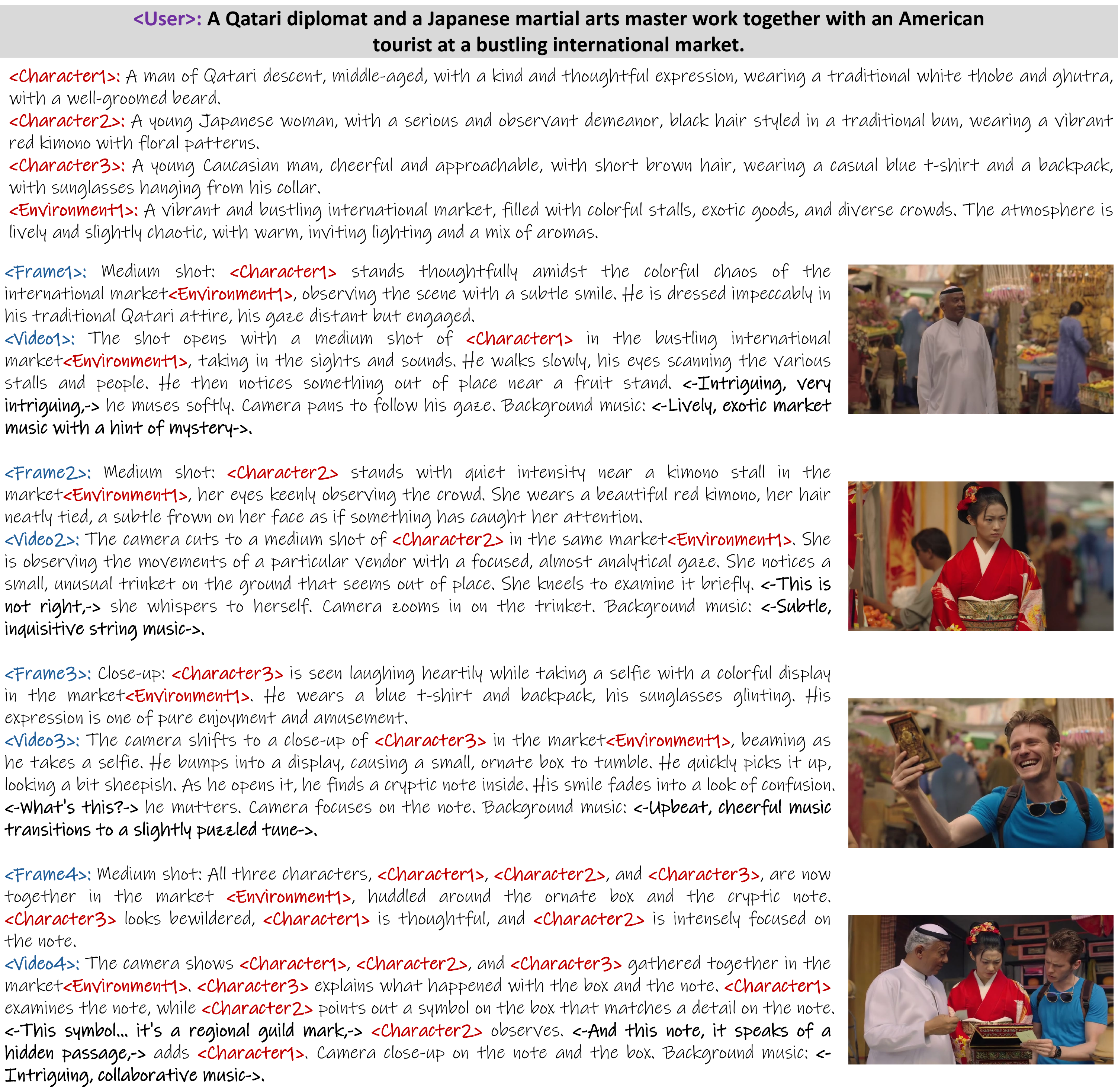}
\end{center}
\label{fig:sup_exp8}
\end{figure}

\begin{figure}[t]
\begin{center}
   \includegraphics[width=1.0\linewidth]{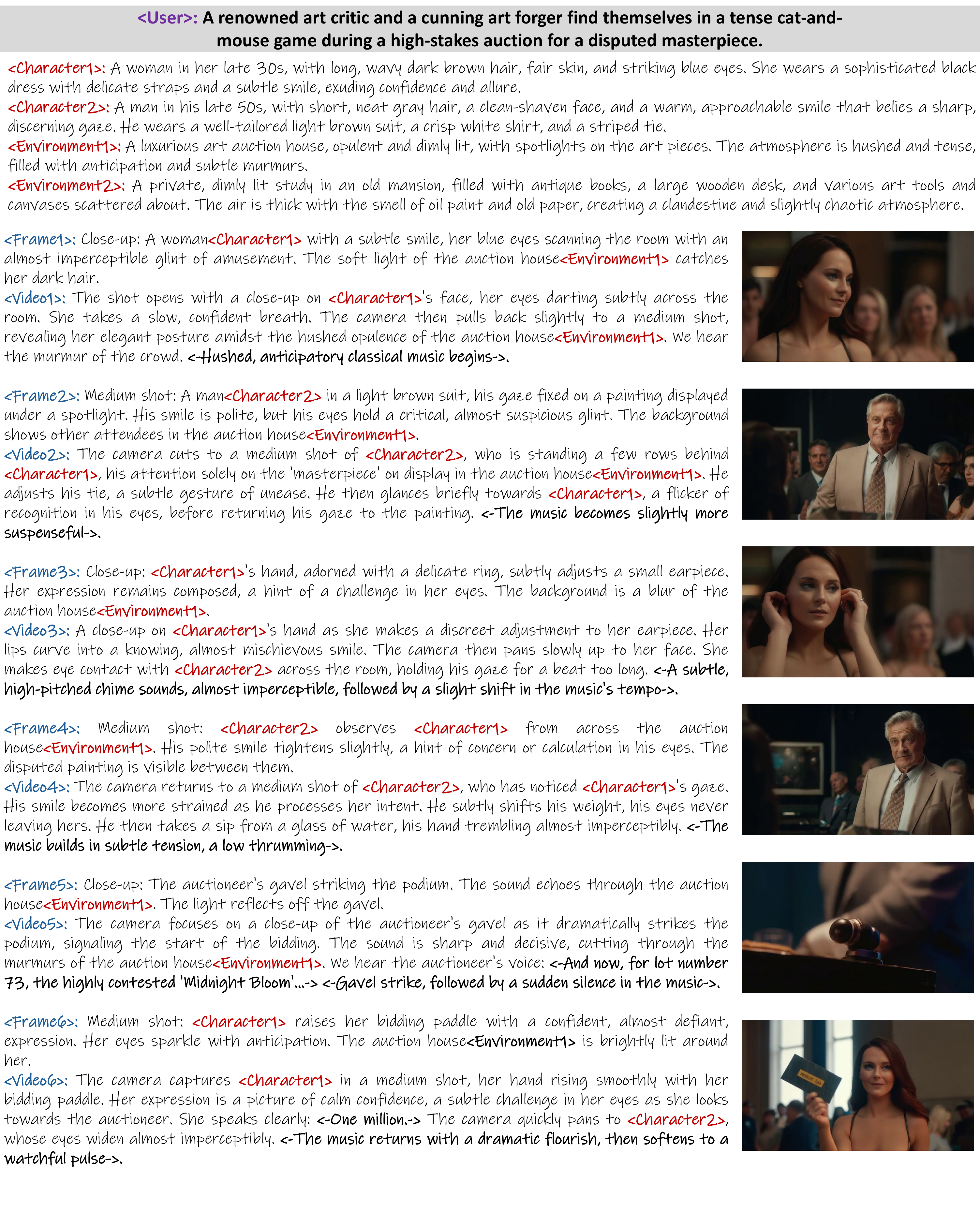}
\end{center}
\label{fig:sup_exp9}
\end{figure}

\begin{figure}[t]
\begin{center}
   \includegraphics[width=1.0\linewidth]{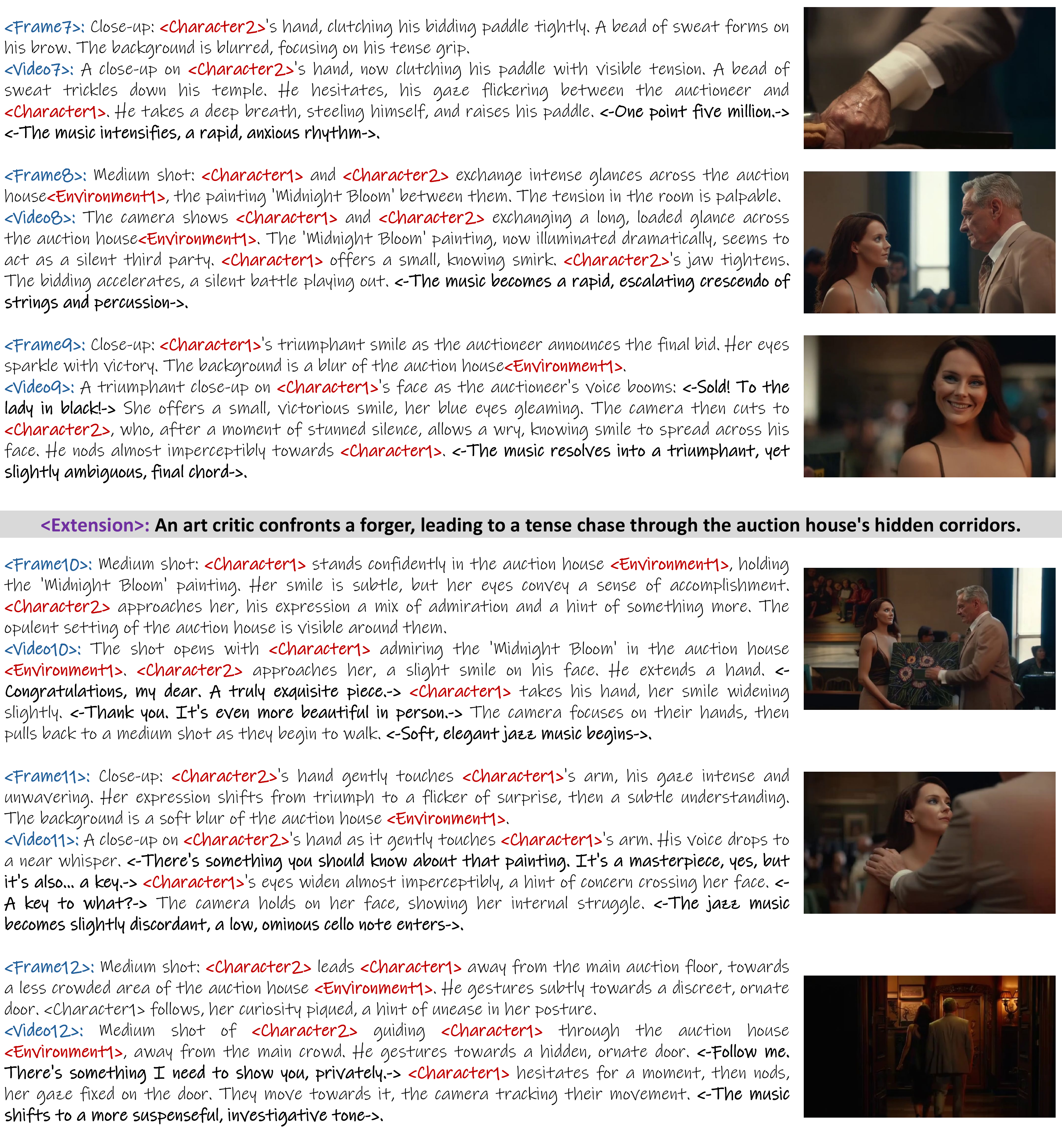}
\end{center}
\label{fig:sup_exp10}
\end{figure}

\begin{figure}[t]
\begin{center}
   \includegraphics[width=1.0\linewidth]{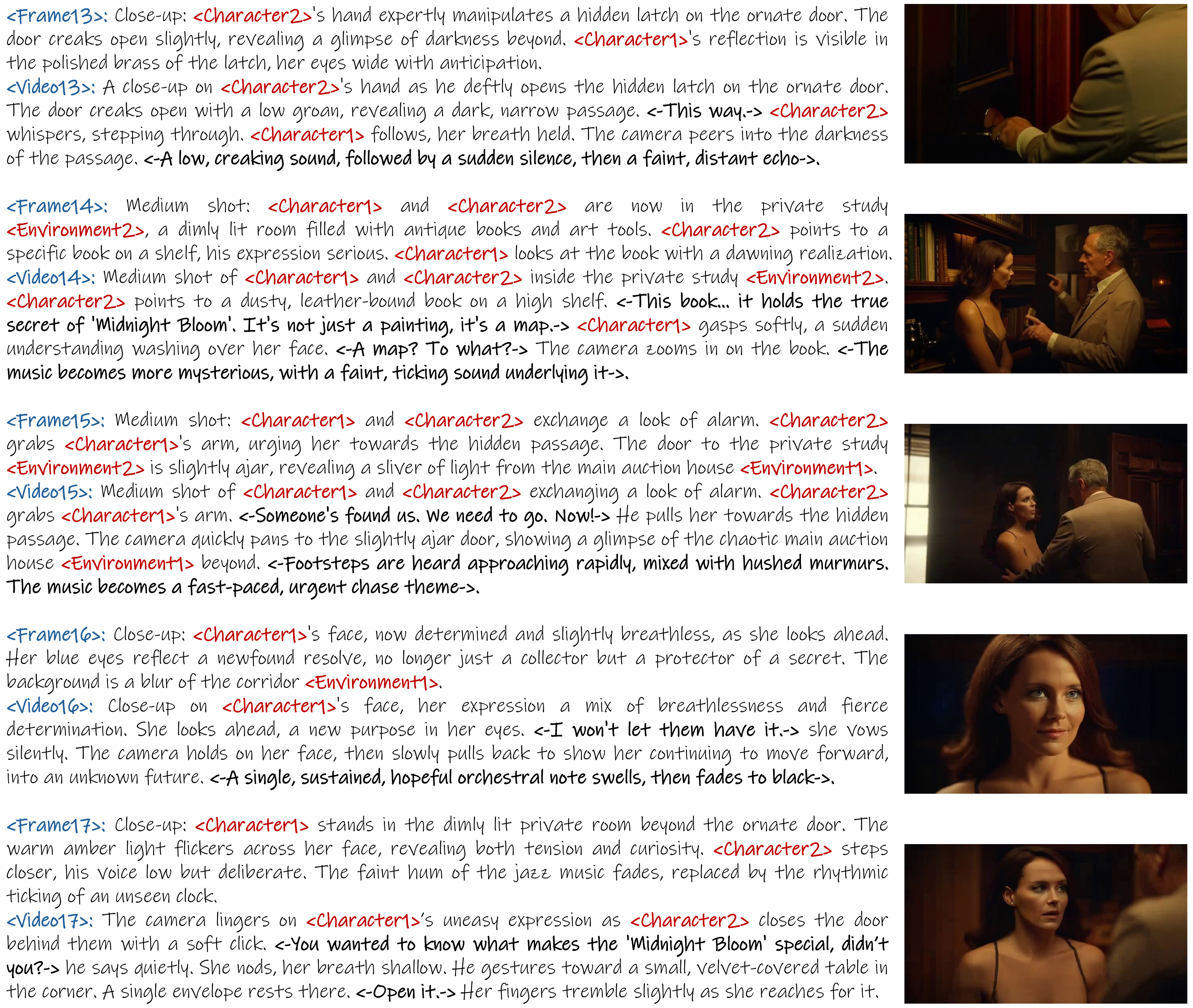}
\end{center}
\label{fig:sup_exp11}
\end{figure}

\begin{figure}[t]
\begin{center}
   \includegraphics[width=1.0\linewidth]{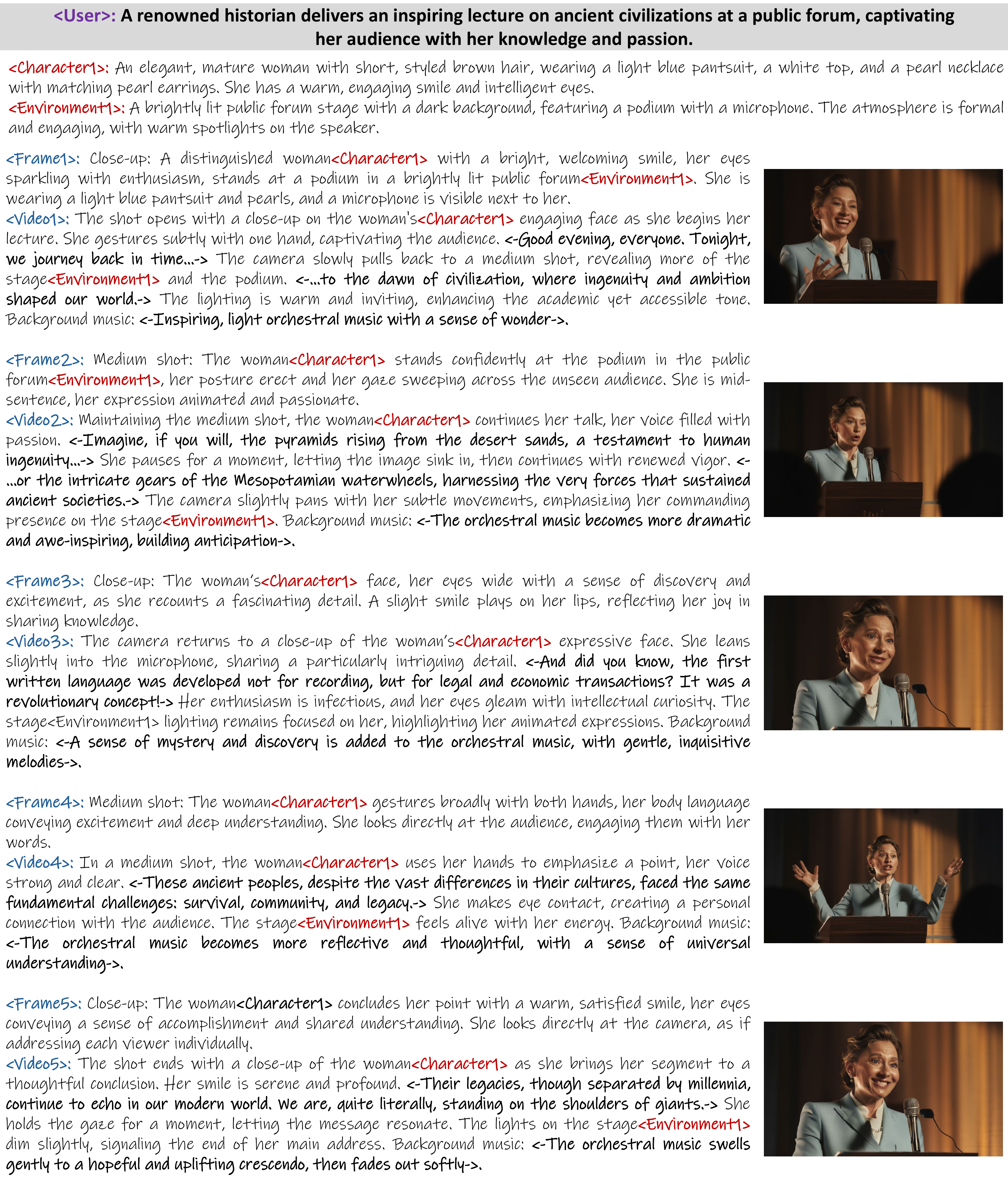}
\end{center}
\label{fig:sup_exp13}
\end{figure}

\begin{figure}[t]
\begin{center}
   \includegraphics[width=1.0\linewidth]{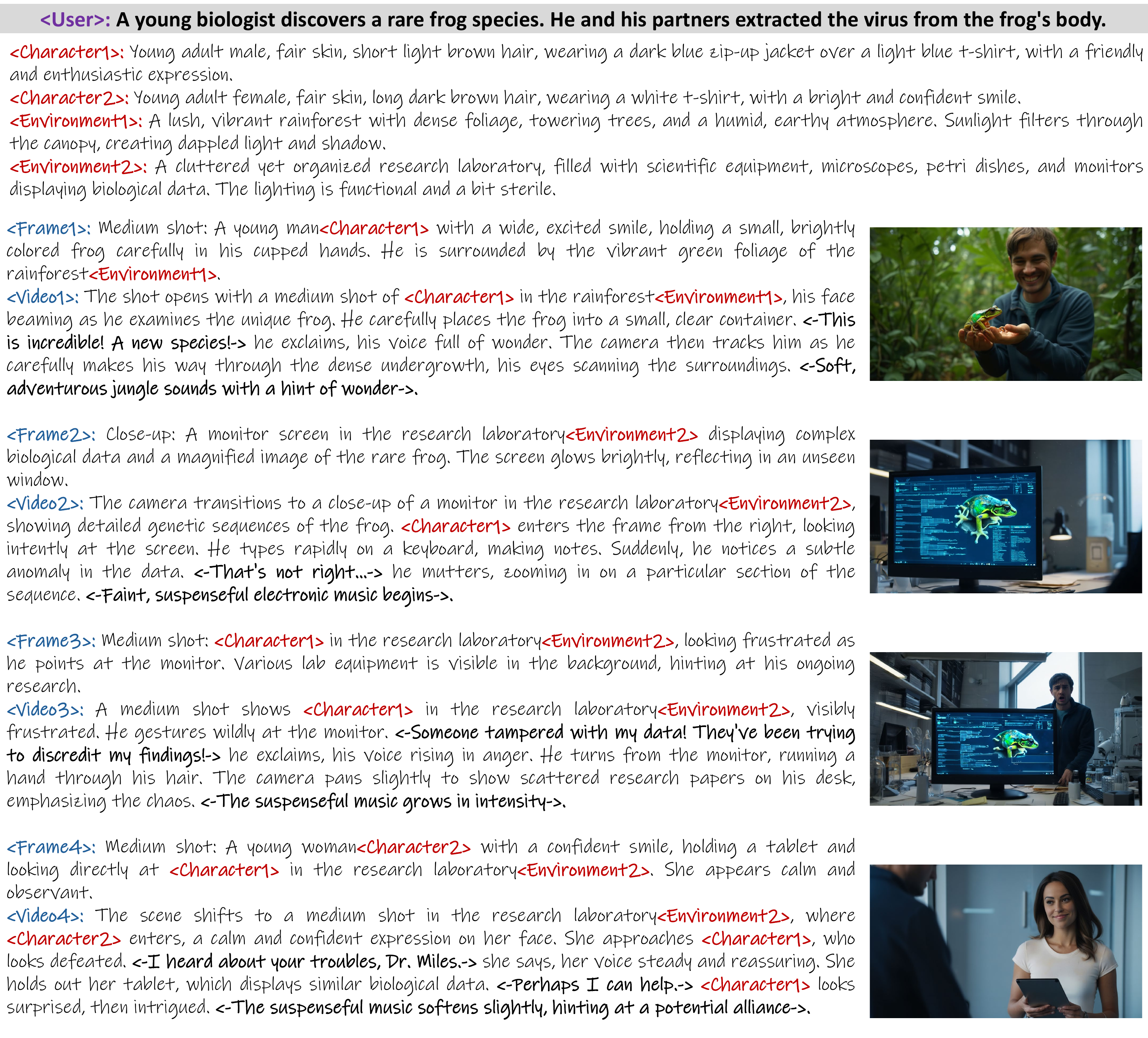}
\end{center}
\label{fig:sup_exp14}
\end{figure}

\begin{figure}[t]
\begin{center}
   \includegraphics[width=1.0\linewidth]{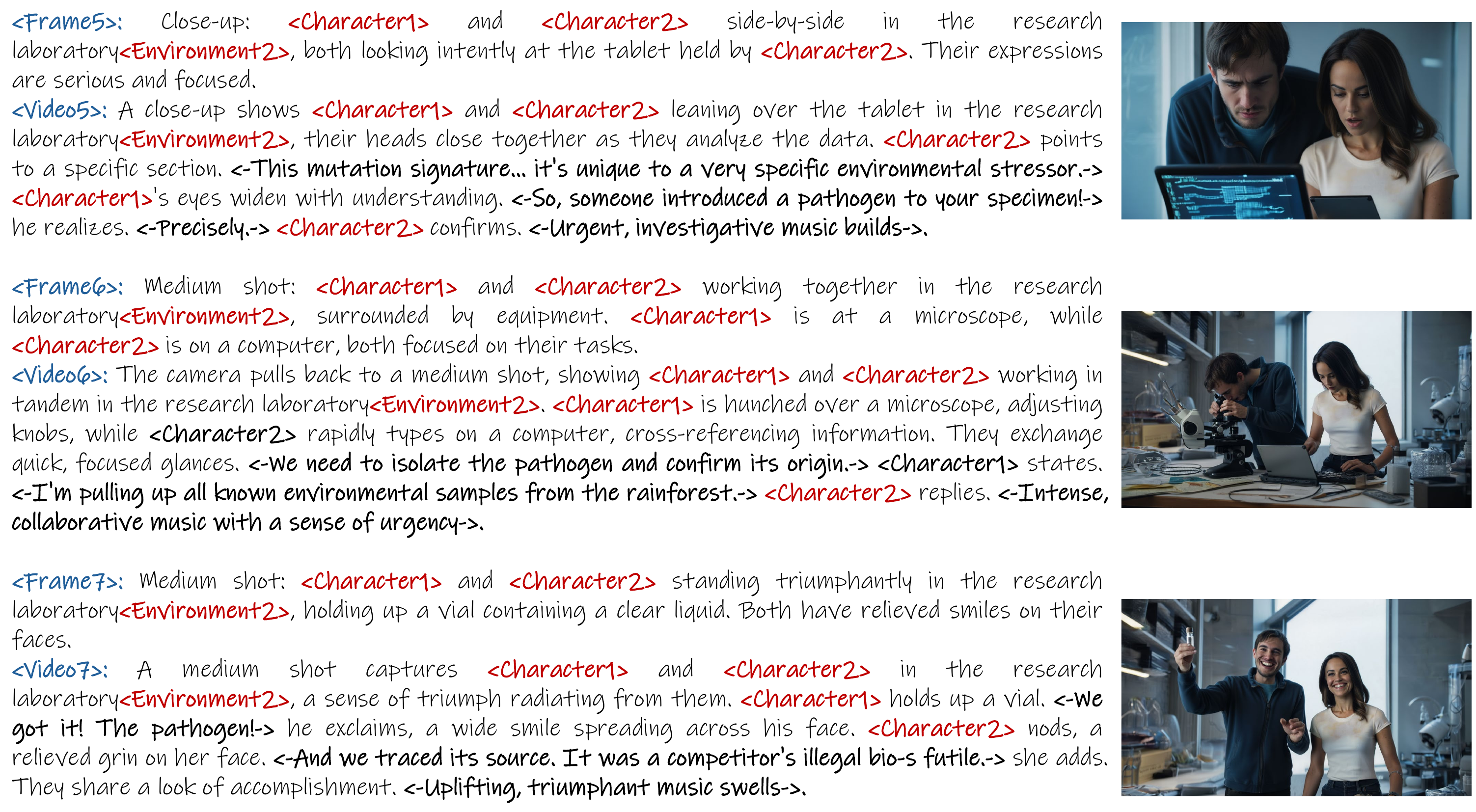}
\end{center}
\label{fig:sup_exp15}
\end{figure}

\begin{figure}[t]
\begin{center}
   \includegraphics[width=1.0\linewidth]{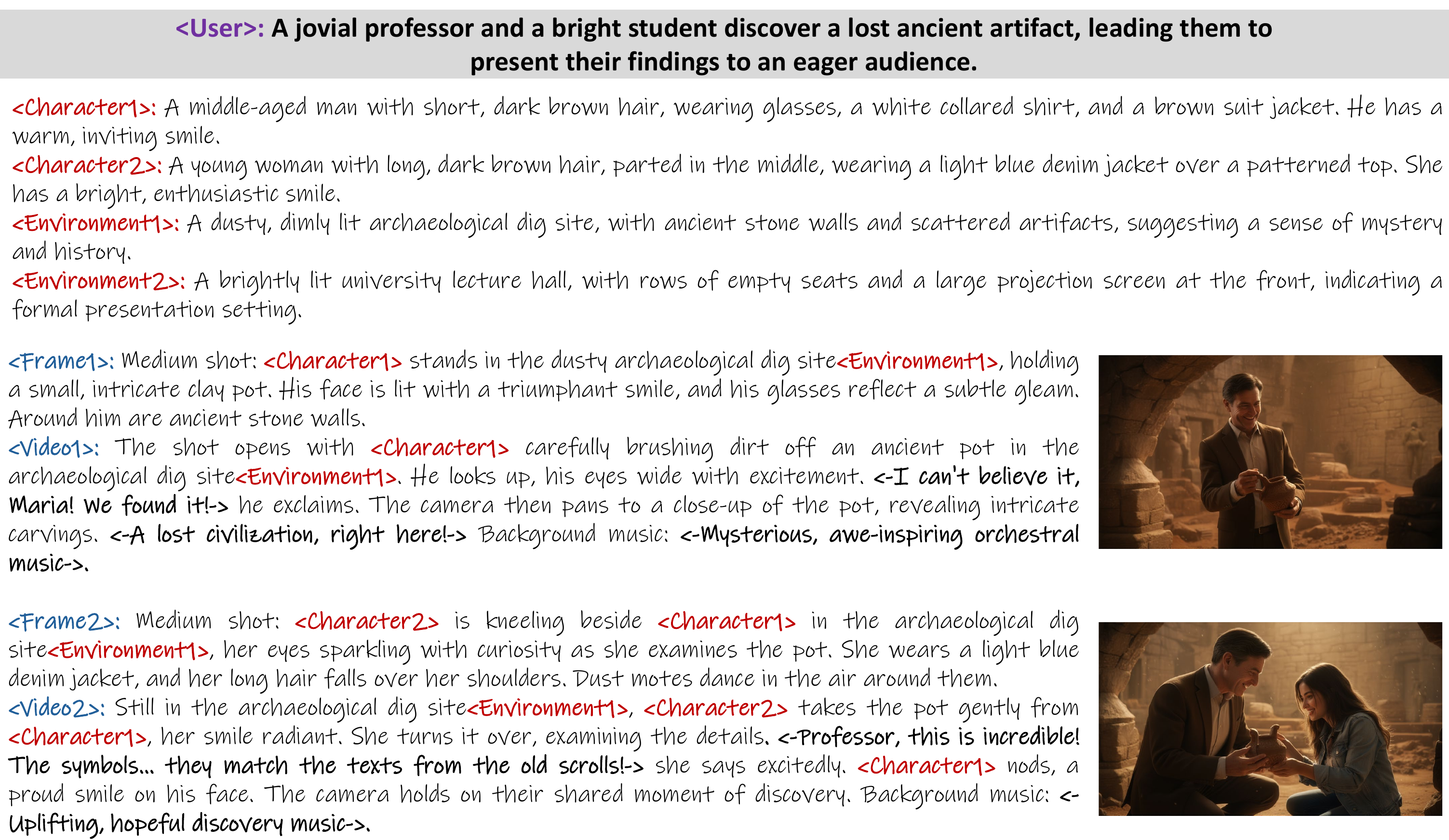}
\end{center}
\label{fig:sup_exp16}
\end{figure}

\begin{figure}[t]
\begin{center}
   \includegraphics[width=1.0\linewidth]{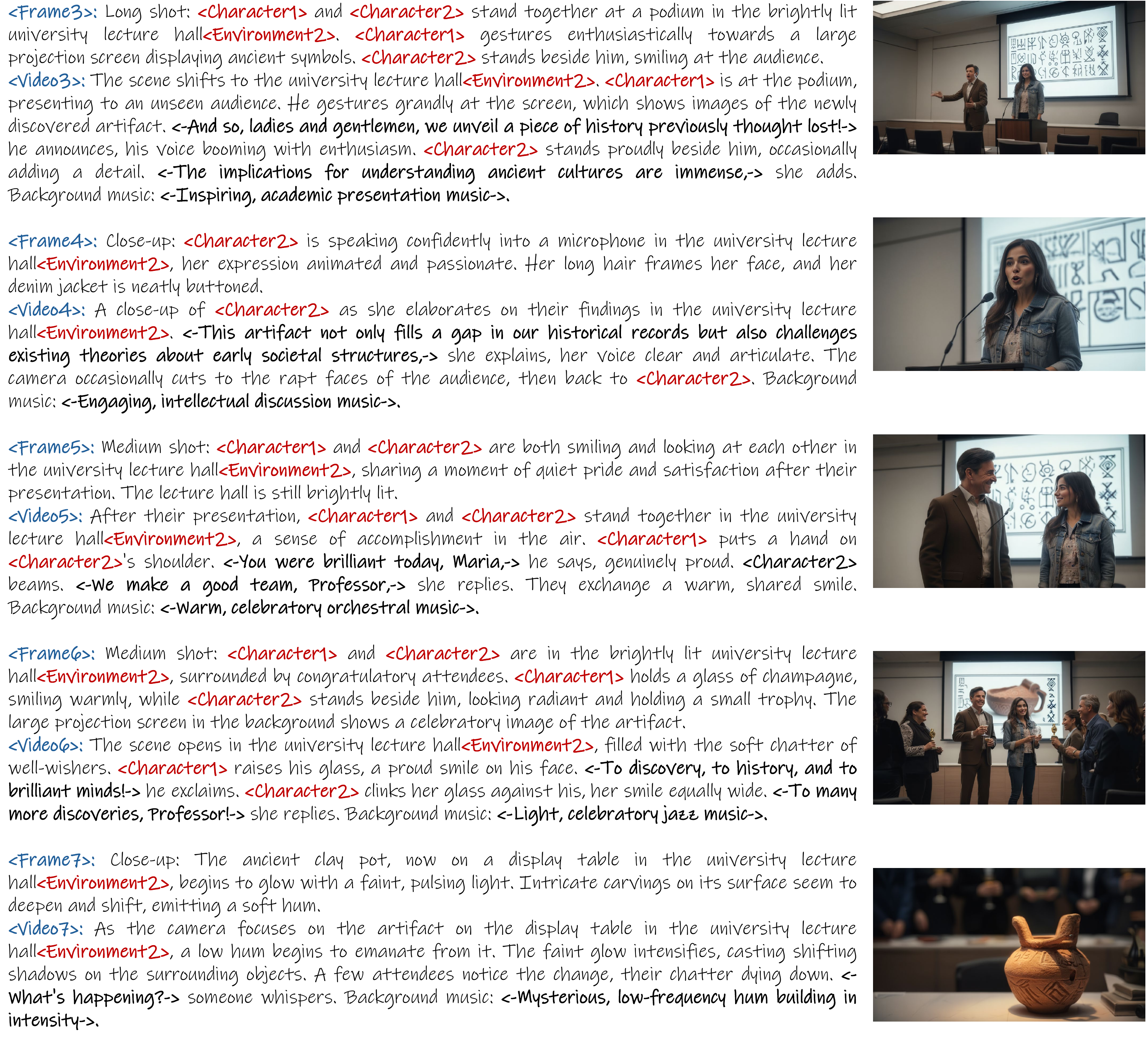}
\end{center}
\label{fig:sup_exp17}
\end{figure}

\begin{figure}[t]
\begin{center}
   \includegraphics[width=1.0\linewidth]{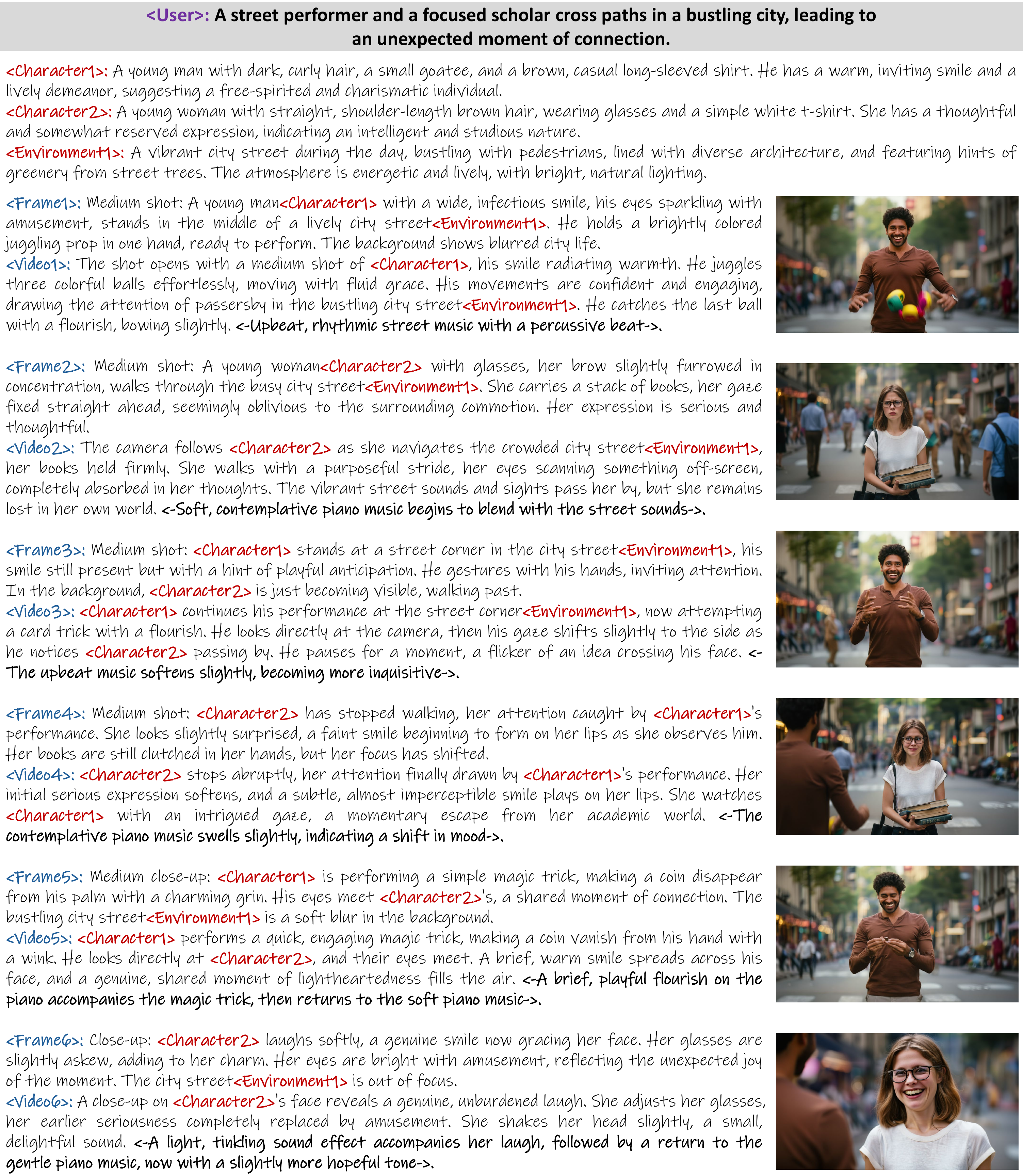}
\end{center}
\label{fig:sup_exp18}
\end{figure}

\begin{figure}[t]
\begin{center}
   \includegraphics[width=1.0\linewidth]{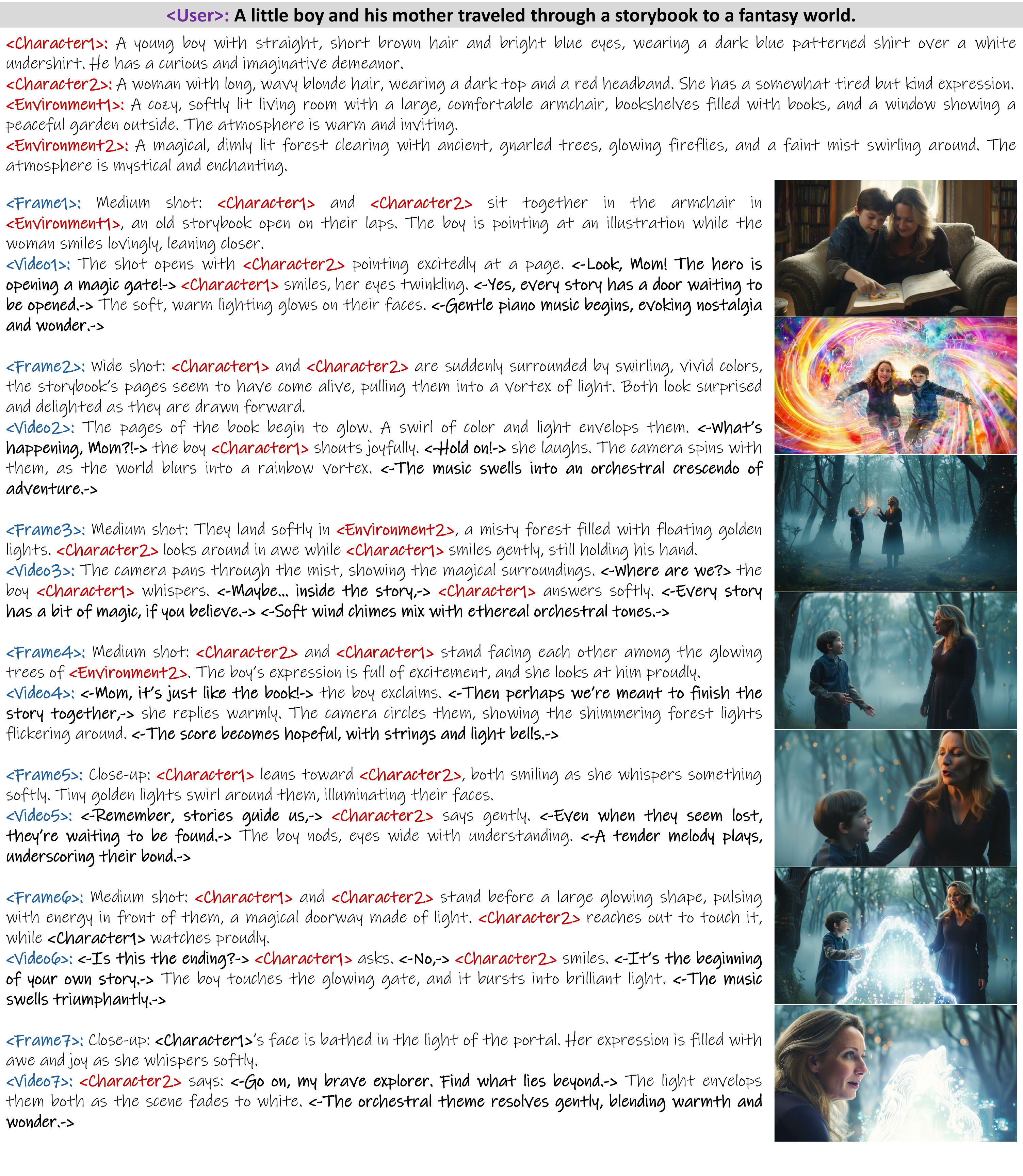}
\end{center}
\label{fig:sup_exp19}
\end{figure}

\begin{figure}[t]
\begin{center}
   \includegraphics[width=1.0\linewidth]{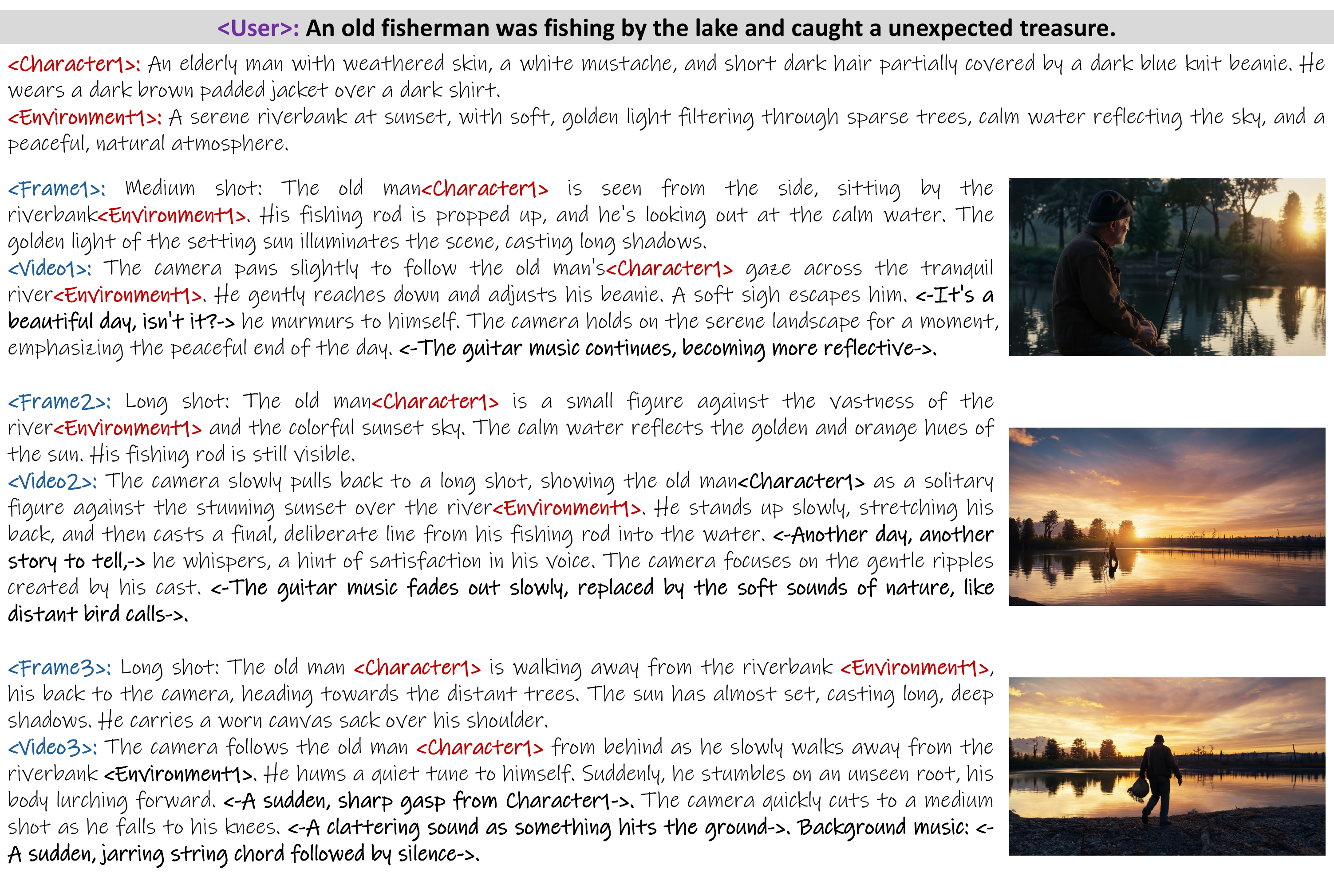}
\end{center}
\label{fig:sup_exp20}
\end{figure}

\begin{figure}[t]
\begin{center}
   \includegraphics[width=1.0\linewidth]{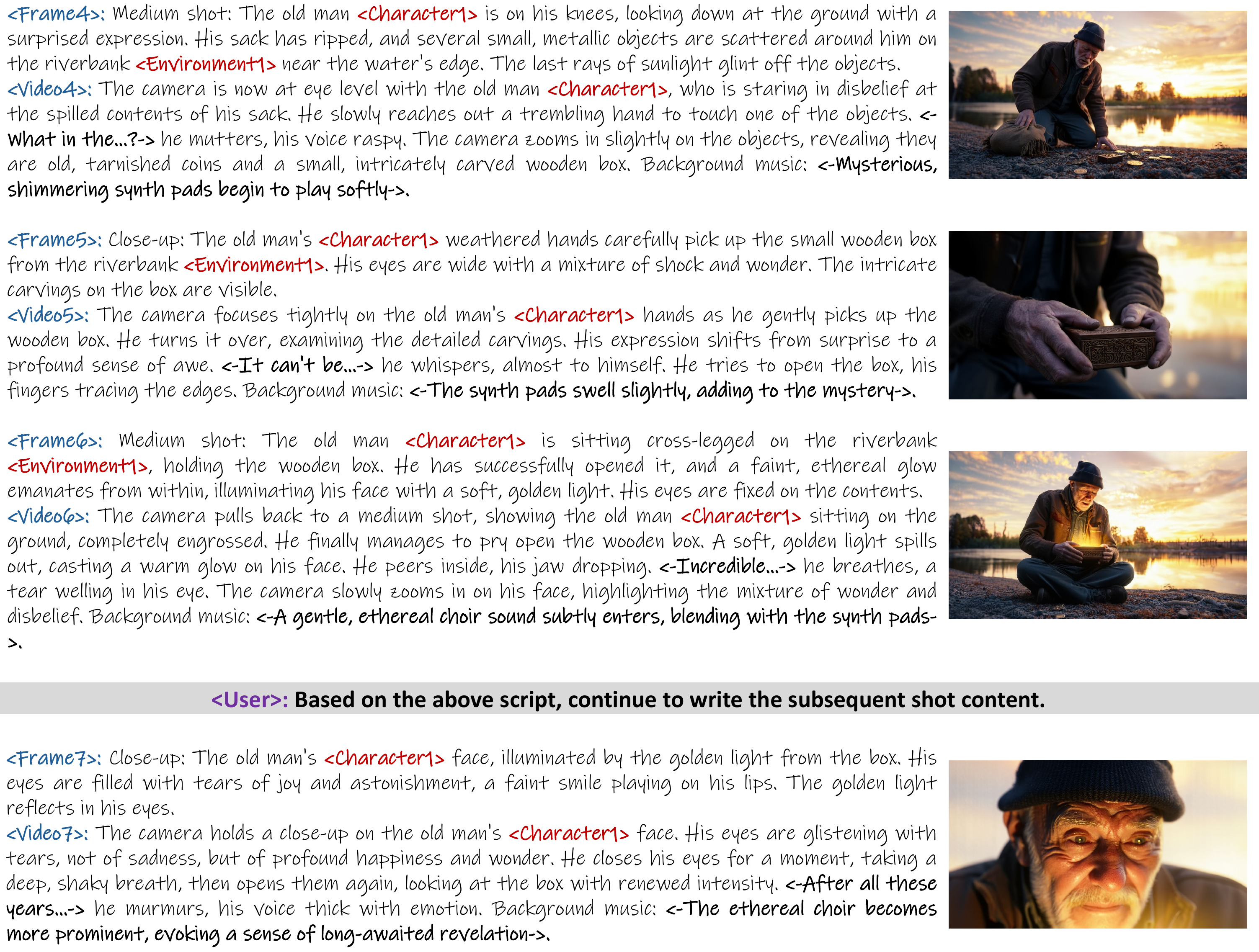}
\end{center}
\label{fig:sup_exp21}
\end{figure}

\begin{figure}[t]
\begin{center}
   \includegraphics[width=1.0\linewidth]{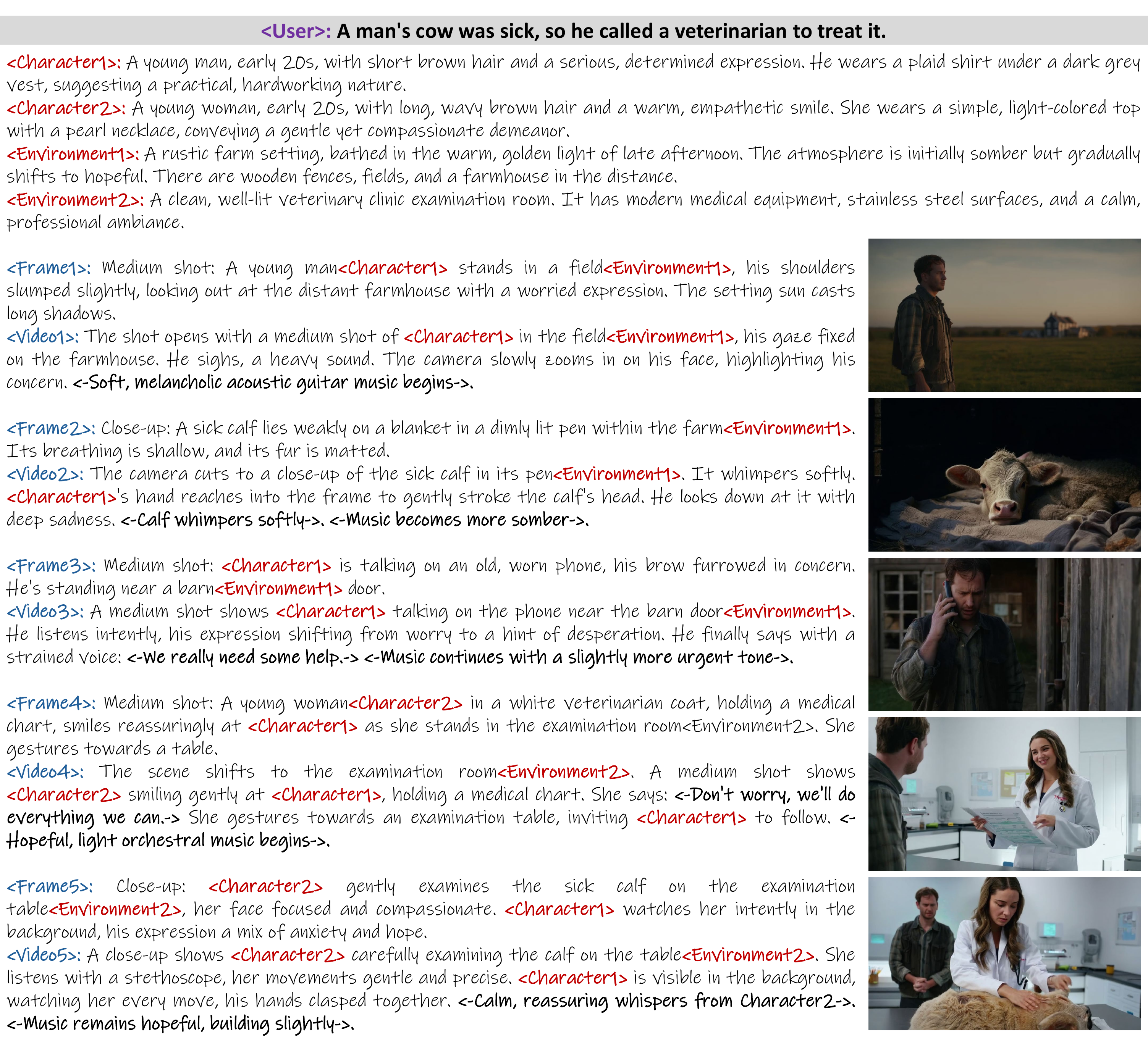}
\end{center}
\label{fig:sup_exp22}
\end{figure}

\begin{figure}[t]
\begin{center}
   \includegraphics[width=1.0\linewidth]{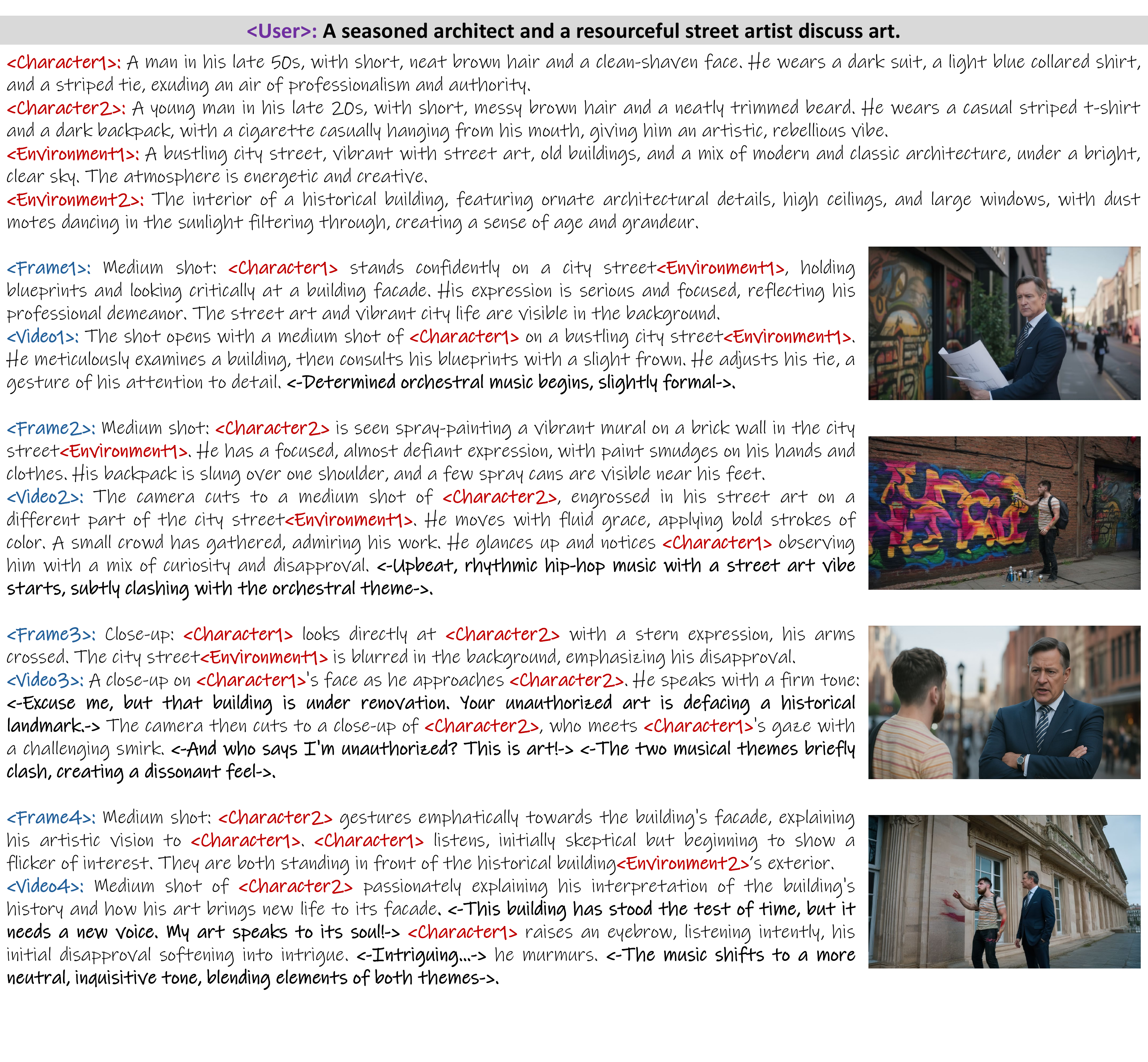}
\end{center}
\label{fig:sup_exp23}
\end{figure}

\begin{figure}[t]
\begin{center}
   \includegraphics[width=1.0\linewidth]{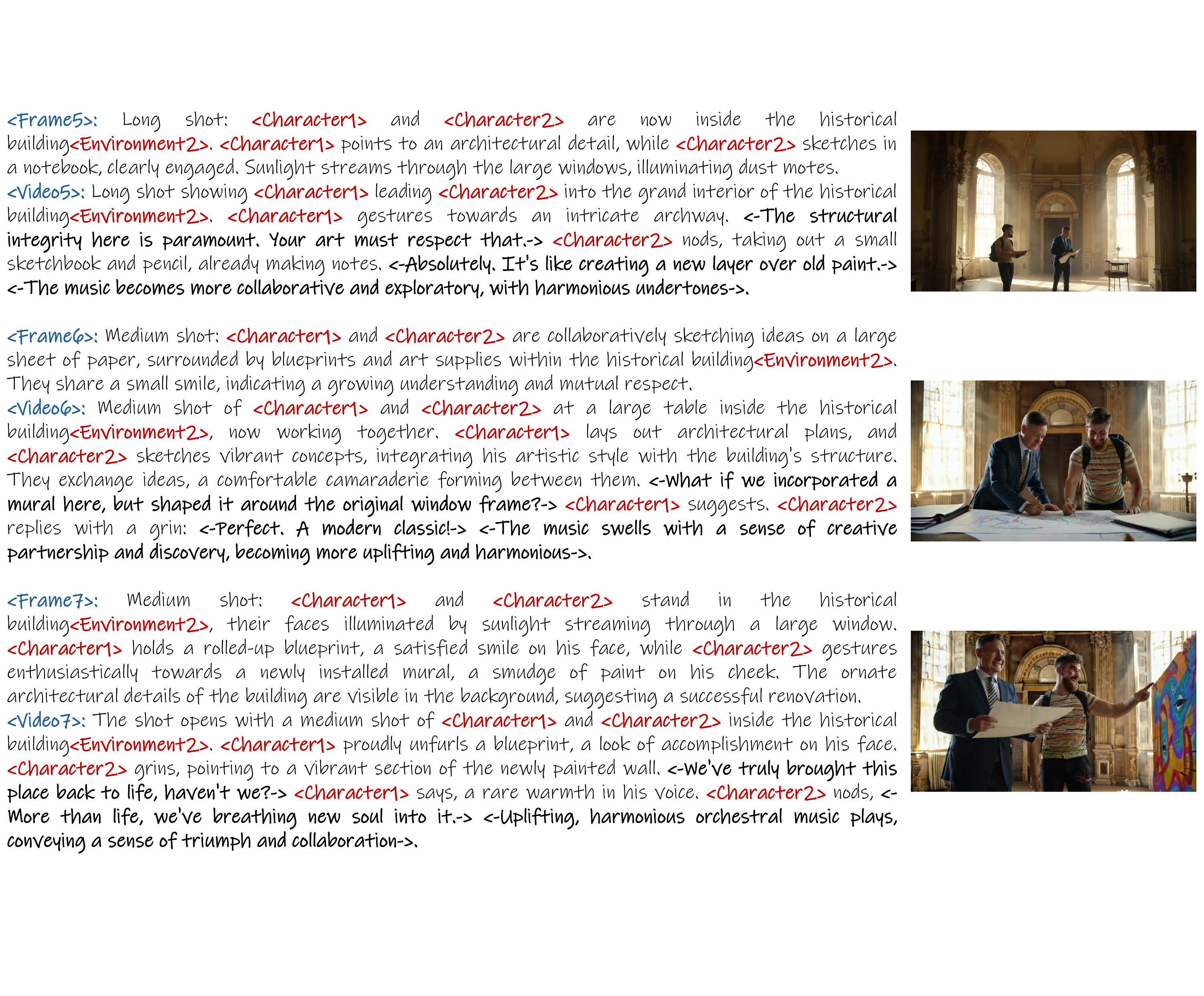}
\end{center}
\label{fig:sup_exp24}
\end{figure}

\begin{figure}[t]
\begin{center}
   \includegraphics[width=1.0\linewidth]{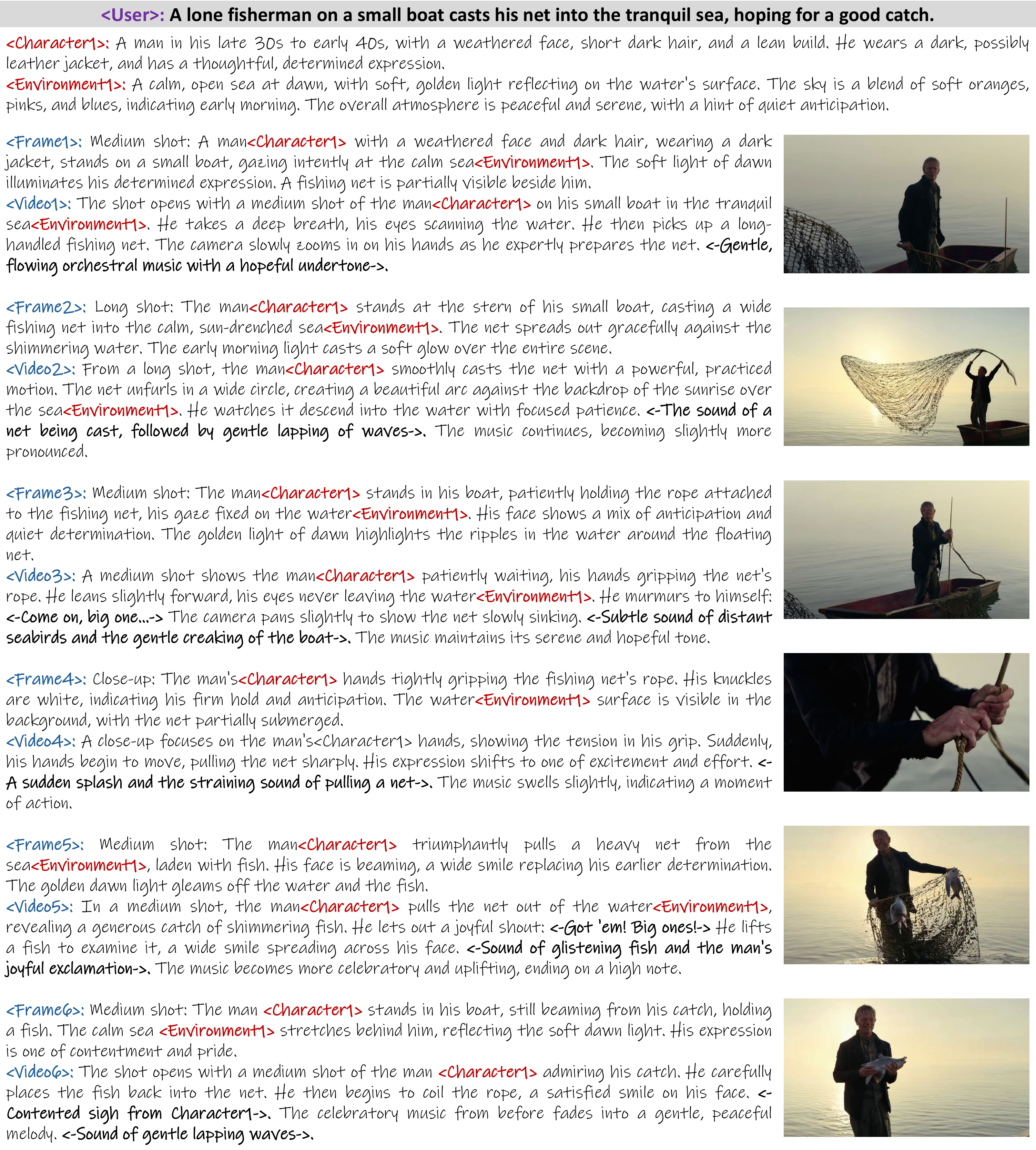}
\end{center}
\label{fig:sup_exp25}
\end{figure}

\begin{figure}[t]
\begin{center}
   \includegraphics[width=1.0\linewidth]{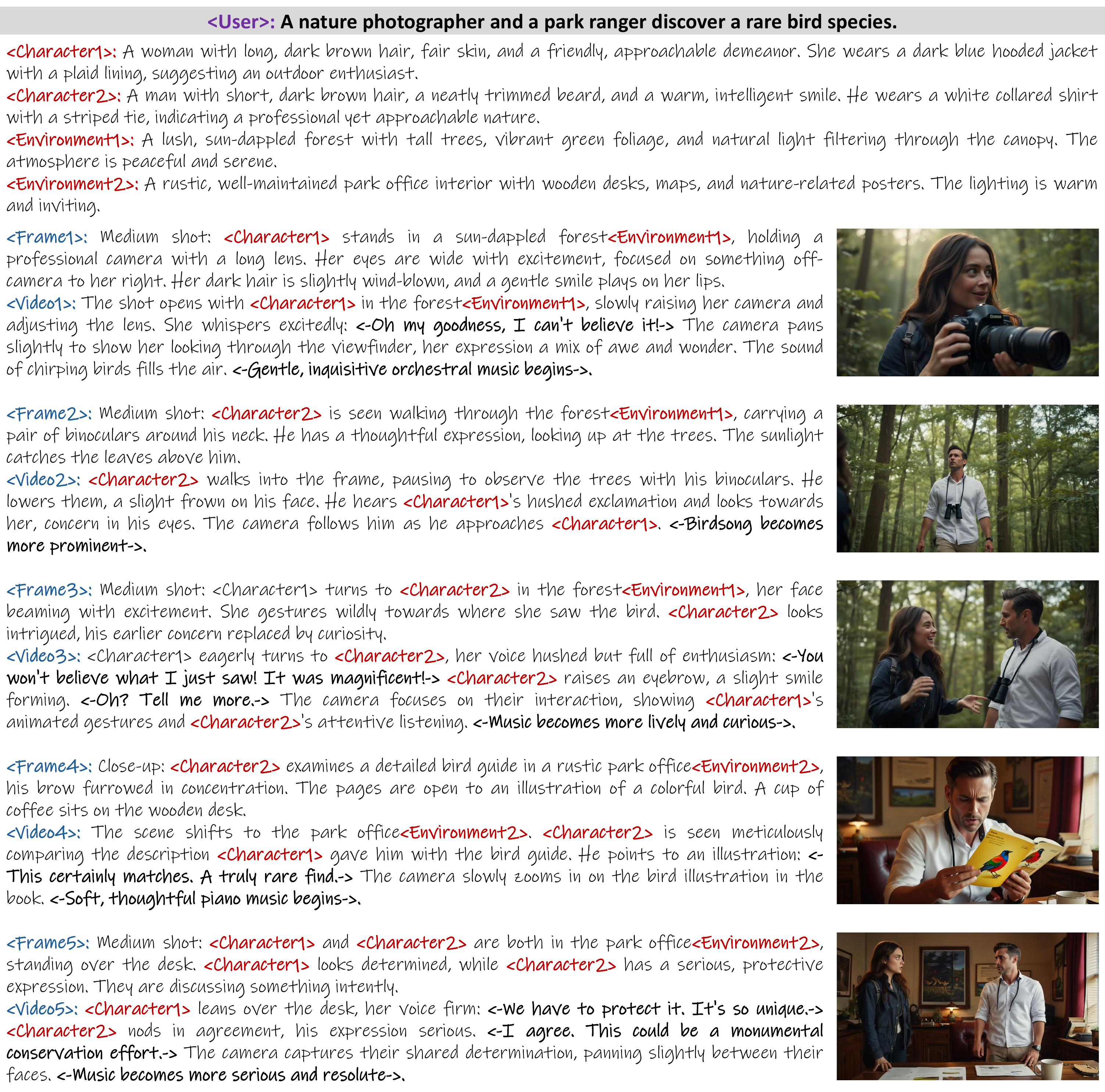}
\end{center}
\label{fig:sup_exp28}
\end{figure}

\begin{figure}[t]
\begin{center}
   \includegraphics[width=1.0\linewidth]{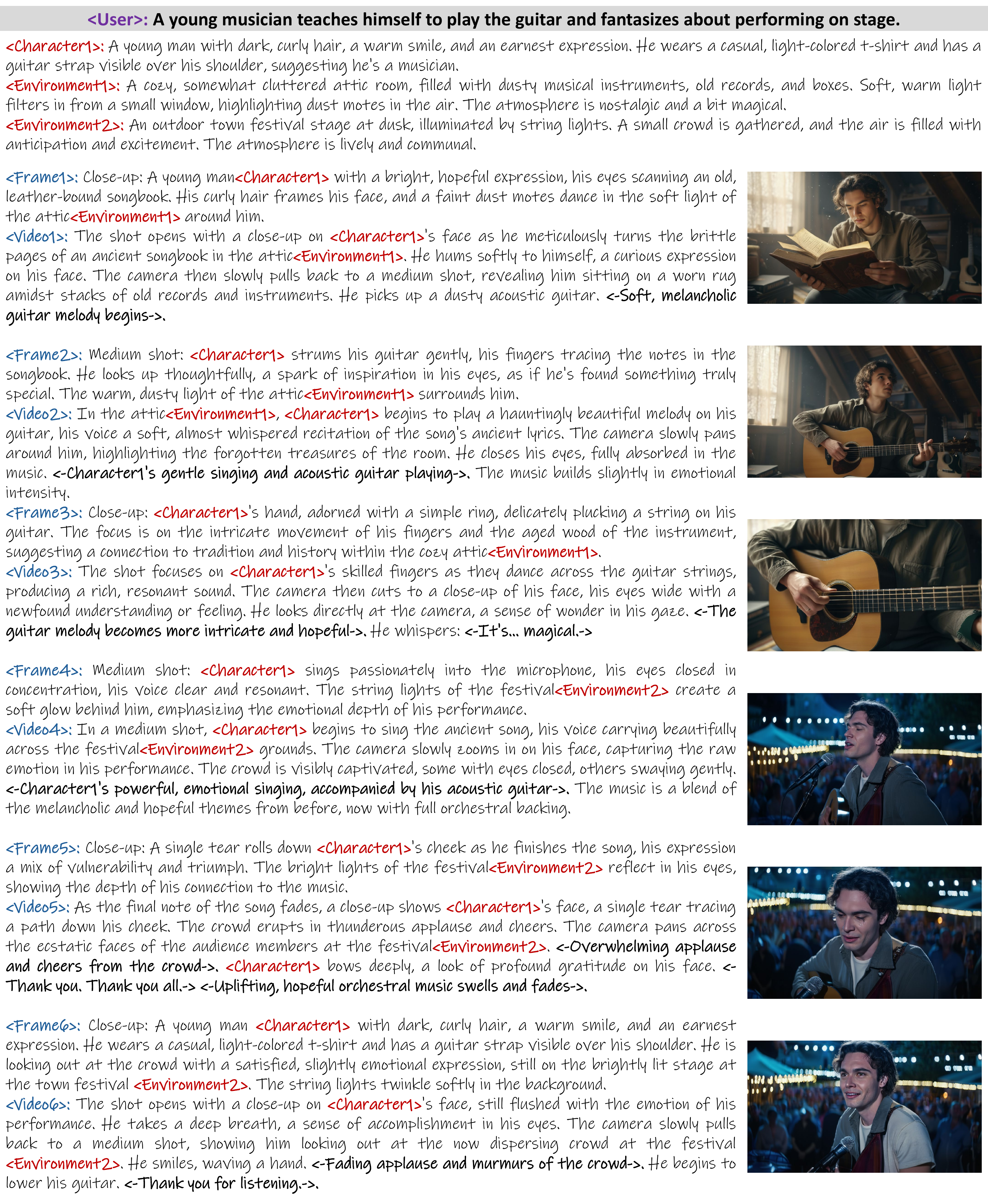}
\end{center}
\label{fig:sup_exp26}
\end{figure}

\end{document}